%% file: is2_freeboard.tex
\documentclass[conference]{IEEEtran}
\IEEEoverridecommandlockouts
\usepackage{cite}
\usepackage{amsmath,amssymb,amsfonts}
\usepackage{algorithmic}
\usepackage{graphicx}
\usepackage{textcomp}
\usepackage{xcolor}

\usepackage{subcaption}
\usepackage{hyperref}
\usepackage{framed}
\usepackage{booktabs}
\usepackage{xcolor}
\usepackage{array}
\usepackage{multirow}
\usepackage{comment}

\usepackage{algorithm}
\usepackage{caption}

\usepackage{authblk} 

\def\BibTeX{{\rm B\kern-.05em{\sc i\kern-.025em b}\kern-.08em
    T\kern-.1667em\lower.7ex\hbox{E}\kern-.125emX}}

\begin{document}

\makeatletter
\newcommand{\linebreakand}{%
  \end{@IEEEauthorhalign}
  \hfill\mbox{}\par
  \mbox{}\hfill\begin{@IEEEauthorhalign}
}
\makeatother

\title{Scalable Higher Resolution Polar Sea Ice Classification and Freeboard Calculation from ICESat-2 ATL03 Data
\\
}
\author{
    Jurdana Masuma Iqrah$^1$, Younghyun Koo$^2$, Wei Wang$^1$, Hongjie Xie$^3$, Sushil K. Prasad$^1$ \\
    $^{1}$Department of Computer Science, University of Texas at San Antonio, San Antonio, TX, USA \\
    $^{2}$National Snow and Ice Data Center (NSIDC), University of Colorado Boulder, Boulder, CO, USA \\
    $^{3}$Department of Earth and Planetary Sciences, University of Texas at San Antonio, San Antonio, TX, USA\\
    Email: jurdanamasuma.iqrah@utsa.edu, younghyun.koo@colorado.edu, wei.wang@utsa.edu,\\ hongjie.xie@utsa.edu, sushil.prasad@utsa.edu

}


\maketitle

\begin{abstract}

ICESat-2 (IS2) by NASA is an Earth-observing satellite that measures high-resolution surface elevation. The IS2's ATL07 and ATL10 sea ice elevation and freeboard products of 10m-200m segments which aggregated 150 signal photons from the raw ATL03 (geolocated photon) data. These aggregated products can potentially overestimate local sea surface height, thus underestimating the calculations of freeboard (sea ice height above sea surface). 
To achieve a higher resolution of sea surface height and freeboard information, in this work we utilize a 2m window to resample the ATL03 data. Then, we classify these 2m segments into thick sea ice, thin ice, and open water using deep learning methods (Long short-term memory and Multi-layer perceptron models). To obtain labeled training data for our deep learning models, we use segmented Sentinel-2 (S2) multi-spectral imagery overlapping with IS2 tracks in space and time to auto-label IS2 data, followed by some manual corrections in the regions of transition between different ice/water types or cloudy regions. We employ a parallel workflow for this auto-labeling using PySpark to scale, and we achieve 9-fold data loading and 16.25-fold map-reduce speedup. To train our models, we employ a Horovod-based distributed deep-learning workflow on a DGX A100 8 GPU cluster, achieving a 7.25-fold speedup.
Next, we calculate the local sea surface heights based on the open water segments. Finally, we scale the freeboard calculation using the derived local sea level and achieve 8.54-fold data loading and 15.7-fold map-reduce speedup. Compared with the ATL07 (local sea level) and ATL10 (freeboard) data products, our results show higher resolutions and accuracy (96.56\%).

\end{abstract}


\section{Introduction}\label{sec:introduction}
\input{sections/introduction}

\section{Related Work}\label{sec:related_work}
\input{sections/related_work}

\section{Methodology}\label{sec:methodology}
\input{sections/methodology}

\section{Experiments and Evaluation}\label{sec:experiments}
\input{sections/experiments}


\section{Conclusion and Future Work}\label{sec:conclusion}
\input{sections/conclusion}




\bibliographystyle{ieeetr}

\bibliography{references/is2_freeboard}

\vspace{12pt}
\color{red}

\end{document}

%% file: sections/introduction.tex

The primary goal of this research is to develop machine learning tools for better processing satellite data for sea ice studies in the polar regions. This inquiry pertains to the profound influence of climate change on GeoScience and society at large, specifically focusing on its impact due to global warming and its consequential effects on ice retreat and melt in the global cryosphere.

The Ross Sea, situated in Antarctica, is a prominent harbor within the Southern Ocean that is renowned for its exceptional and pristine nature. Although the discussion surrounding the weather's impact on global warming tends to focus on other regions, it is crucial to consider the changes occurring in the Ross Sea. These changes offer valuable insights into the broader implications of climate change and warming trends. The examination of the Ross Sea and its associated meteorological patterns can provide valuable insights into the comprehension of global warming. In general, the documented alterations in the Ross Sea, encompassing its sea ice, temperatures, and ecosystems, offer significant insights into the intricate dynamics between regional and worldwide meteorological patterns and their association with the phenomenon of global warming. The examination of these locations facilitates a better comprehension of the Earth's climate system and its reaction to alterations caused by human activities.

NASA's ICESat-2 (Ice, Cloud, and Land Elevation Satellite-2) mission is to measure the elevation of Earth's surface, especially its ice sheets, sea ice, and vegetation. One of the key datasets provided by ICESat-2 (IS2) is the ATL03 product that contains precise measurements of the height of Earth's surface, along with geolocation and other information \cite{neumann2019ice}. 
A multitude of Earth science topics, including climate change, polar ice sheet dynamics, and vegetation monitoring, are studied using ATL03 data and other higher-level products. Through NASA's Earthdata Search and other data distribution platforms, the dataset is readily accessible to the public, allowing scientists and researchers worldwide to access and analyze the data for their studies and applications.

The ATL03 data includes the height, latitude, longitude, geolocated photon elevation, and time of individual photons, and it is a large dataset. On the other hand, ATL07 and ATL10 are additional ICESat-2 data products that are derived from ATL03 and measure sea ice height and sea ice freeboard, respectively. 
Freeboard is the thickness of sea ice protruding above the water level. 
These ATL07 and ATL10, level 3 data products are derived from ~150 signal photon aggregation of ATL03, a level 2 data product \cite{kwok2020icesat}.
The ATL07 product comprises along-the-track segments of sea surface and open water leads (at varying length scales) height relative to the WGS84 ellipsoid (ITRF2014 reference frame) after adjustment for geoidal and tidal variations and inverted barometer effects. The along-track length of these segments depends on the distance over which ~150 signal photons (of ATL03) are accumulated; as a result, it can vary depending on the surface type \cite{kwok2020icesat}. The ATL10 product consists of the sea-ice freeboard calculated, each within swath segments that are 10 km (nominally) along the track and 6 km (the distance between the six beams) across the track. For freeboard calculation, the segments of the freeboard swath are utilized to construct a reference sea surface. 
The ATL07 and ATL10 products are accumulations of 150 signal photons of ATL03 and have 10m-200m spatial resolution for strong beams and 20m-400m for weak beams \cite{kwok2019surface}. However, the ATL03 data, which has a resolution of 11m footprint with 0.7m spacing, is too big to store and process for domain sea ice scientists due to its huge volume of data. In this study, we adopt a 2m sampling strategy to reprocess the ATL03 data, and we use distributed computing and deep learning technology for processing and classifying the resampled ATL03 data into thick ice, thin ice, and open water. We then derive a higher resolution of local sea level and freeboard products. We aim to get better-resolution products to achieve better scientific sea ice dynamics results from this ATL03 data than the ATL07 and ATL10.

To classify and calculate sea ice surface height and freeboard retrieval, NASA used a decision tree-based approach \cite{kwok2020icesat} on ATL07 data. Nonetheless, this product has the weakness of having a lower resolution than the ATL03.
We propose to use deep learning approaches, namely Multi-layer Perceptron (MLP) and Long Short Term Memory (LSTM), to achieve better sea ice classification accuracy for the ATL03 data. 
Labeled data are required for training to apply deep learning-based approaches for sea ice classification. To label the ATL03 data, we first selected correlated Sentinel-2 (S2) \cite{drusch2012sentinel} images within an 80-minute temporal extent between IS2 and S2. These S2 sea ice images were auto-labeled based on our thin-cloud and shadow-filtered color-based segmentation method\cite{iqrah2023toward}. With the labeled S2 images, we can then map/overlay them to the correlated ALT03 data and automatically transfer the S2 labels to label the thick ice, thin ice and open water in IS2 ATL03 track line data. 
To handle the large amount of ATL03 data labeling, we utilize distributed parallel computing. We use data parallelism and distributed deep learning, utilizing the Horovod framework \cite{sergeev2018horovod} to scale and speedup the deep learning training over multiple GPU machines without sacrificing classification accuracy.

Our parallel workflow includes distributed computing for auto-labeling and freeboard computation, as well as distributed deep-learning training. The distributed scaled auto-labeling of IS2 data achieved around 16.25x speedup, and distributed freeboard computation achieved similar around 15.7x speedup. The distributed LSTM-based sea ice classification model achieved 96.56\% than the MLP model with 91.84\% classification accuracy, achieving a 7.25x speedup on an 8 GPU DGX cluster.

The following are the primary contributions of this paper:
\begin{itemize}
    \item ATL03 sea ice and open water labeled training data using correlated S2 imagery,
    \item Deep learning-based (LSTM and MLP) sea ice classification,
    \item Higher resolution local sea level and freeboard information retrieval using 2m sampling of ATL03 dataset and sea surface estimation based on open water class.
    \item Our distributed computing framework achieved a 16.25x speedup for auto-labeling and 15.70x for freeboard computation, while distributed deep learning training achieved a 7.25x speedup on 8 GPUs.
\end{itemize}


The remaining sections of the paper are organized as follows. Section 2 reviews the key related work. Section 3 describes our proposed methodology for sea ice classification and freeboard retrieval. Section 4 contains the evaluation metrics, experimental results, and a discussion of the proposed methodologies. Finally, in Section 5, we provide concluding remarks and suggest future directions for this ongoing work.

%% file: sections/related_work.tex


NASA's 
ICESat-2 can detect sea ice features due to its high spatial resolution data products. For example, \cite{farrell2020mapping} used ICESat-2 ATL03 geolocated photon data to retrieve six dynamic properties of sea ice, including surface roughness, ridge height, ridge frequency, melt pond depth, floe size distribution, and lead frequency.
Another work on IS2 ATL03 \cite{fredensborg2020estimation} also presented that the degree of ice ridging can be retrieved from this data precisely.

To examine the surface classification \cite{petty2021assessment}, they utilized ICESat-2 ATL07 and ATL10 sea ice products using near-coincident optical images from S2 over the Western Weddell Sea in March 2019 and the Lincoln Sea in May 2019. However, S2 overlays suggest cloud-induced dark lead misdiagnosis. As a result, they need adjustments to select sea surface reference points \cite{kwok2021refining}. 
Apart from that, in general, this decision-tree-based approach of the ATL07 product shows a good performance on the sea ice and open water classification when compared with other high-resolution satellite images \cite{kwok2019surface}, \cite{kwok2021refining}, \cite{petty2021assessment}.
This paper \cite{xu2021deriving}, proposed an improved One-Layer Method (OLMi) for Antarctic sea-ice thickness retrieval with an uncertainty of 0.3 m on ICESat (IS) and IS2. This method examines IS2's monthly sea ice variance and thickness in the Antarctic, demonstrating bi-modal distributions. They also estimate freeboard consistency between IS and IS2.
An initial study \cite{kwok2020arctic} to compare satellite lidar (ICESat-2) and radar (CryoSat-2) freeboards to estimate Arctic sea ice snow depth. They determined that the sea ice thickness can be calculated with snow loading from satellite retrievals without resorting to climatology or reconstructions.
In \cite{koo2023sea}, for sea ice surface type classification, they utilized coincident S2 to manually label the ATL07 data into different sea ice surface types (thick/snow-covered ice, thin ice, and open water) for building and validating machine learning models. The validated MLP model (99\% accuracy) was used to classify sea ice surface types and then used to derive freeboard. Additionally, in \cite{koo2021weekly}, provided a weekly mapping of freeboard and analysis for the Ross Sea, Antarctic, using the IS2 ATL10 freeboard products.

This \cite{ball2017comprehensive} presented a thorough survey of environmental remote sensing and deep learning research here. They also concentrated on unsolved challenges and opportunities related to (i) inadequate data sets, (ii) human-understandable solutions for modeling physical phenomena, (iii) big data, (iv) nontraditional heterogeneous data sources, (v) DL architectures and learning algorithms for spectral, spatial, and temporal data, (vi) transfer learning, (vii) an improved theoretical understanding of DL systems, (viii) high barriers to entry, and (ix) training and optimizing.
This review \cite{yuan2020deep} proceeded into a detailed discussion of the potential for deep learning in the analysis and prediction of environmental remote sensing data. They also assessed deep-learning environmental monitoring for surface temperature, atmosphere, evapotranspiration, hydrology, vegetation, etc.

Therefore, this study represents a novel endeavor to apply machine learning techniques to IS2 ATL03 data to classify sea ice cover types. This research aims to develop innovative machine-learning models for ATL03 surface classification, aiming to achieve better resolution and accuracy in determining sea ice classification and freeboard. Ultimately, these advancements in sea ice classification will contribute to a deeper understanding of sea ice dynamics in polar regions.

%% file: sections/methodology.tex
\begin{figure}[!ht]
    \centering
    \frame{\includegraphics[width=1.0\linewidth]{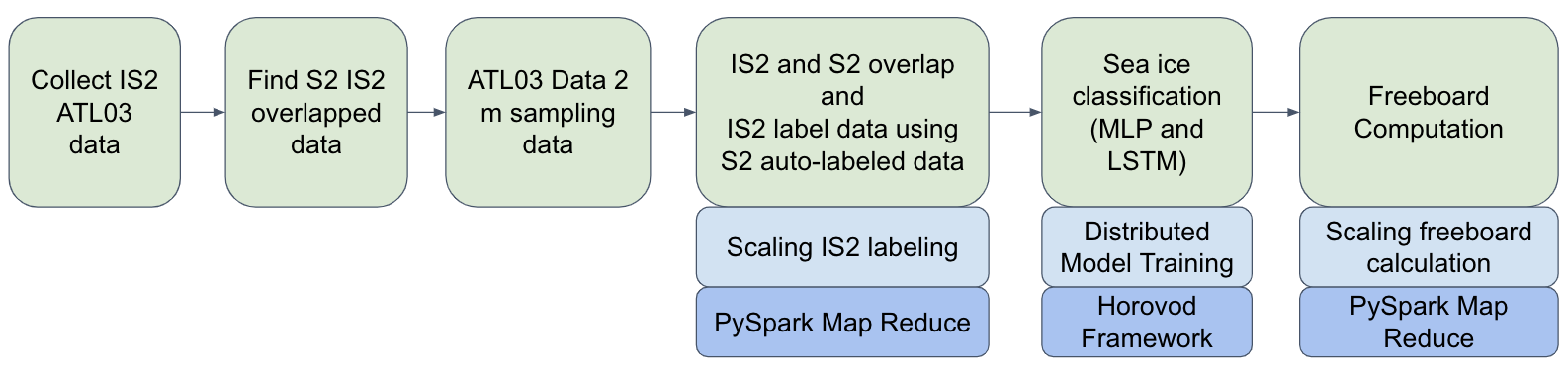}}
\caption{ATL03 Sea Ice Classification and Freeboard Computation Workflow}
\label{fig:is2_workflow}
\end{figure}

This study focuses on the sea ice classification and freeboard retrieval workflow in the Antarctic Ross Sea region. This workflow comprises four primary stages. In the first stage, data collection, processing, and auto-labeling are conducted to prepare labeled data for training and scaling the auto-labeling process. Following that, we perform a distributed training of deep learning models, encompassing Multi-layer Perceptron (MLP) \cite{riedmiller2014multi} and Long Short-Term Memory (LSTM) \cite{hochreiter1997long} networks. The third stage involves scaled and distributed model inference to obtain sea ice classification data. In the fourth stage, distributed local sea level detection and computation of freeboard are performed. Figure \ref{fig:is2_workflow} demonstrates the ATL03 sea ice classification and Freeboard Computation Workflow.

\subsection{Data Curation}

\subsubsection{Region of Interest}
The Ross Sea, located in a deep embayment of the Southern Ocean, holds the title of the southernmost sea on Earth.
Strong katabatic winds in the Ross Sea dynamically push sea ice away from the coast and ice shelves on a daily or weekly basis \cite{bromwich1993hemispheric}, \cite{dale2017atmospheric}, creating open polynyas (areas of open water or newly formed thin ice) \cite{bromwich1993hemispheric}, \cite{thompson2020frazil}. Three significant and persistent polynyas identified in the Ross Sea are the Ross Ice Shelf Polynya, the Terra Nova Bay Polynya, and the McMurdo Sound Polynya. The interplay of strong katabatic winds and polynya formation is vital for understanding the behavior of sea ice in the Ross Sea \cite{dai2020ice}, \cite{kwok2007ross}, \cite{tian2020sea}.
Hence, in this study, the Ross Sea region is used as an experimental site to demonstrate the development of the sea ice dynamics algorithms and workflow. 
The spatial extent of the Ross Sea is from longitude -180 to -140 and latitude -78 to -70.

\begin{figure}[htb]
\begin{framed}
        \centering
        \begin{subfigure}[b]{0.32\textwidth} 
            \centering
            \includegraphics[width=\textwidth]{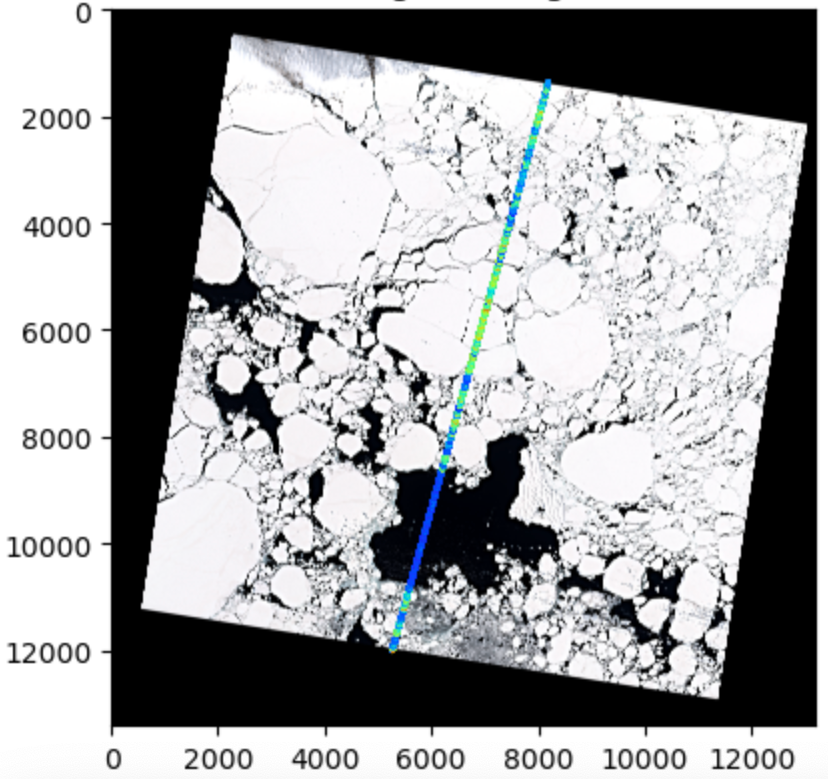}
            \caption{}
            \label{fig:autolabeling_ori_elev}
        \end{subfigure}
        \begin{subfigure}[b]{0.32\textwidth}
            \centering
            \includegraphics[width=\textwidth]{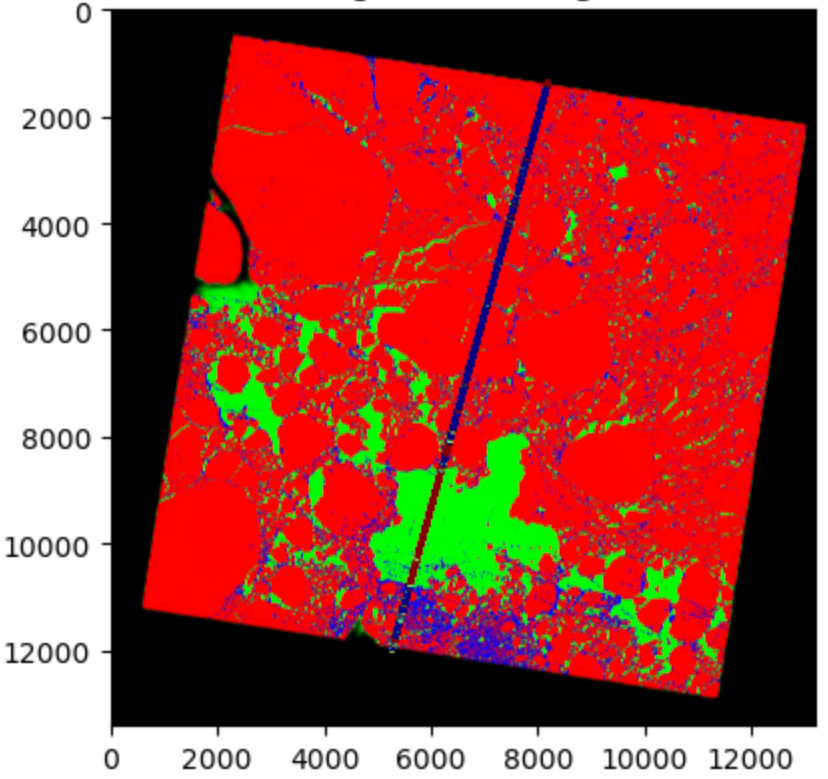}
            \caption{}
            \label{fig:autolabeling_seg}
        \end{subfigure}
        \begin{subfigure}[b]{0.32\textwidth}
            \centering
            \includegraphics[width=\textwidth]{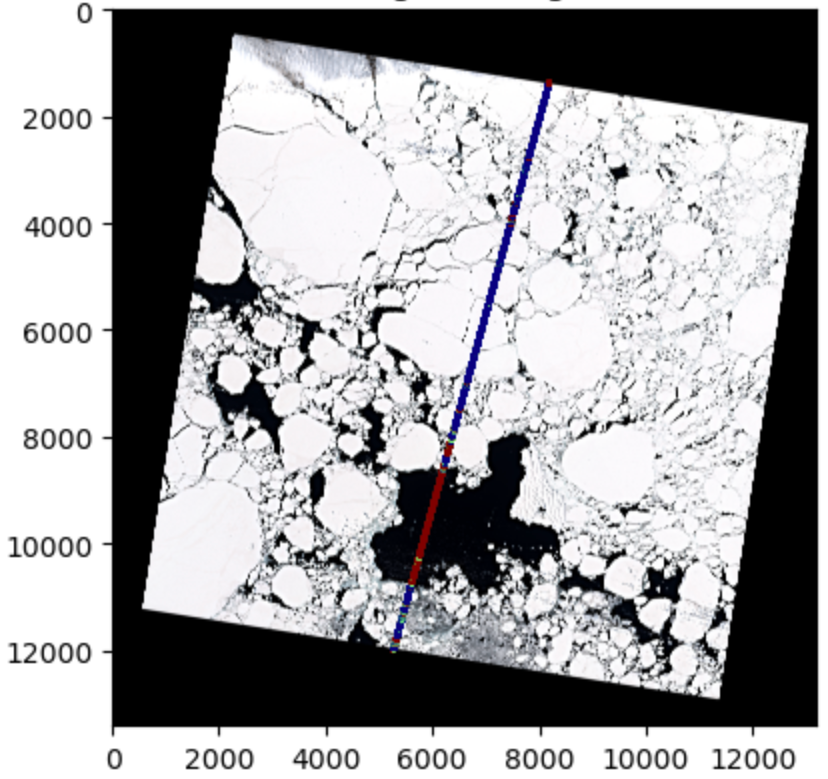}
            \caption{}
            \label{fig:autolabeling_ori}
        \end{subfigure}
    \end{framed}
    \caption{Auto-labeling of IS2 (line) elevations into thick sea ice, thin ice, and open water based on S2 (image) classified surface types: (a) IS2 track elevation over S2 image, (b) Auto-labeling IS2 surface types using S2 classified surface types, and (c) Auto-labeled IS2 surface types over S2 image.}
    \label{fig:autolabeling}
\end{figure}

\subsubsection{Data Preprocessing}
This research is based on the ATL03 (release 006) sea ice geolocated photon data over the Ross Sea region. The data are obtained from the NASA Earthdata server. Only the three strong beams of the ATL03 tracks are used for the study. For each of the beams, we collect geolocation elevation along with other parameters based on the signal classification confidence of the high sea ice surface type. We also calculate the background factors, apply a geographical correction based on \cite{neumann2021icesat}, and remove ineffective reference photons. After that, we sample ATL03 data for every 2m to calculate statistical parameters, such as mean, median, and standard deviation for height, elevation, photon count, background photon, etc. Finally, we apply the first-photon bias correction to the data to maintain its correctness.
\subsubsection{Auto-Label Data}
Since sea ice classification requires labeled ATL03 data, we select correlated Sentinel-2 (S2) imagery for labeling. First, we search for ATL03 track line data from the Ross Sea region in November 2019. Then, we collect correlated S2 images for the same region with up to an 80-minute temporal window difference between these two satellite data. We apply thin cloud and shadow-filtered color-based segmentation \cite{iqrah2023toward} to auto-label the S2 imagery. This technique for labeling can handle thin clouds and thin shadows as satellites are often affected by the atmospheric cloud and shadow cover. Then we use the labeled data and overlay the ATL03 data on it using the same geographical projection code (EPSG 3976) for both datasets. This same projection code is needed to compare the IS2 data points with the S2 data points. After that, we auto-label the IS2 ATL03 data point based on the corresponding S2 image sea ice labels. A sample autolabeling of the IS2 track line using a 10m resolution S2 image is given in the figure \ref{fig:autolabeling}. 
%
%
However, there is a temporal difference between IS2 and S2 datasets, as shown in Table 1. Due to this time difference, in some cases, there is a drift in the sea ice, causing the overlapped S2 and IS2 data to be misaligned. To correct this misalignment, we need to calculate the drift and adjust the IS2 and S2 data positions to get better alignment. Therefore, we shift the coincident S2 images to align them with the IS2 track based on the sea ice (thick ice, thin ice, and open water) labels from the S2 images and the elevation value of the IS2 ATL03 data product. The shift of S2 images is accomplished based on the Table \ref{shift_table}. Since these alignments are not perfect, the transition regions of different types of sea ice are affecting the correctness of the data labeling. We also found that due to some thick cloud and shadow cover in the S2 images, the overlapped IS2 track data are mislabeled. To handle these mislabeling issues resulting from the auto-labeling process, we manually correct the cloudy and transition regions to obtain a better labeled training data for IS2 data.

\begin{table}[htb]

\caption{IS2 ATL03 and S2 coincident pairs ($<2$ h of time difference) in the Ross Sea in November 2019. S2 images are shifted to match the IS2 data.}
\label{shift_table}
\begin{tabular}{|c|c|c|c|c|}
\hline
  & \begin{tabular}[c]{@{}c@{}}IS2 \\ acquisition time\\ (UTC)\end{tabular} & \begin{tabular}[c]{@{}c@{}}S2 \\ acquisition time\\ (UTC)\end{tabular} & \begin{tabular}[c]{@{}c@{}}Time \\ difference \\ (minutes)\end{tabular} & \begin{tabular}[c]{@{}c@{}}Shift of \\ S2 images\\ (distance /\\  direction)\end{tabular} \\ \hline
1 & \begin{tabular}[c]{@{}c@{}}2019/11/03\\ 18:44:32\end{tabular}           & \begin{tabular}[c]{@{}c@{}}2019/11/03\\ 18:34:59\end{tabular}          & 9.55                                                                    & 550 m / NW                                                                                \\ \hline
2 & \begin{tabular}[c]{@{}c@{}}2019/11/04\\ 19:53:11\end{tabular}           & \begin{tabular}[c]{@{}c@{}}2019/11/04\\ 19:45:29\end{tabular}          & 7.7                                                                     & 0 m                                                                                       \\ \hline
3 & \begin{tabular}[c]{@{}c@{}}2019/11/13\\ 19:10:53\end{tabular}           & \begin{tabular}[c]{@{}c@{}}2019/11/13\\ 18:34:59\end{tabular}          & 35.9                                                                    & 200 m / W                                                                                 \\ \hline
4 & \begin{tabular}[c]{@{}c@{}}2019/11/16\\ 19:28:13\end{tabular}           & \begin{tabular}[c]{@{}c@{}}2019/11/16\\ 18:44:59\end{tabular}          & 43.23                                                                   & 0 m                                                                                       \\ \hline
5 & \begin{tabular}[c]{@{}c@{}}2019/11/17\\ 19:02:34\end{tabular}           & \begin{tabular}[c]{@{}c@{}}2019/11/17\\ 18:15:09\end{tabular}          & 47.57                                                                   & 530 m / NW                                                                                \\ \hline
6 & \begin{tabular}[c]{@{}c@{}}2019/11/20\\ 19:19:52\end{tabular}           & \begin{tabular}[c]{@{}c@{}}2019/11/20\\ 20:05:29\end{tabular}          & 45.62                                                                   & 400 m / NW                                                                                \\ \hline
7 & \begin{tabular}[c]{@{}c@{}}2019/11/23\\ 18:02:55\end{tabular}           & \begin{tabular}[c]{@{}c@{}}2019/11/23\\ 18:34:59\end{tabular}          & 32.07                                                                   & 150 m / E                                                                                 \\ \hline
8 & \begin{tabular}[c]{@{}c@{}}2019/11/26\\ 18:20:14\end{tabular}           & \begin{tabular}[c]{@{}c@{}}2019/11/26\\ 18:44:59\end{tabular}          & 24.75                                                                   & 350 m / SW                                                                                \\ \hline
\end{tabular}
\end{table}

\subsubsection{Scaled IS2 Auto-labeling}
First, to scale the finding of IS2 and S2 overlapped data and labeling of the IS2 based on corresponding S2 data, we used the pyspark-based map-reduce framework. Our labeling process is highly data-parallel.
Given the substantial volume of IS2 data, partitioning the dataset and distributing the computations across multiple machines is essential for improving scalability and expediting processing. To facilitate this, we leverage the PySpark map-reduce framework. This PySpark framework enables effective partitioning of large datasets and distribution of the workload across multiple worker nodes, thus enabling data parallelism and enhancing the efficiency of our project.

\subsection{Deep Learning Model Training}
For the sea ice classification problem, we explore two techniques: LSTM and MLP. The MLP model is generally more useful for classification problems. However, the LSTM model performs better in terms of sensor-based data.

For sea ice classification in ATL07 from ATL03, NASA used their own decision tree-based approach. However, this product has the disadvantage of having a lower resolution than the original ATL03. The decision tree is a relatively simple machine-learning technique utilized for both classification and regression purposes. 
Nevertheless, the decision tree algorithm is susceptible to overfitting, particularly when the tree is deep and intricate, rendering it ill-suited for capturing complicated patterns within extensive datasets.
As ICESat-2 data has lots of complex information with complicated patterns, we decided to employ a Deep Neural Network approach for handling this complex spatial dataset.   


\subsubsection{LSTM}
The LSTM \cite{hochreiter1997long} is a special type of recurrent neural network (RNN) that has been developed to effectively process sequential data, such as time series 
and textual information. LSTM models have the ability to capture and model long-range relationships included in sequential data effectively. This characteristic makes LSTMs particularly suitable for tasks that require the analysis of temporal patterns and context. 
This is achieved by the utilization of memory cells, which enable the capturing of long-range dependencies. Consequently, RNNs have found extensive application, mostly over the time series data classification and prediction. Since ATL03 is time series sensor data, and since we are doing classification on this big data, the LSTM model can help to achieve better results. 
In this study, we aim to classify each data point from high-resolution ATL03 data into three categories, namely i) Thick Ice, ii) Thin Ice, and iii) Open Water. By analyzing our data, we saw that the properties of one point, such as height/elevation, height standard deviation, high-confidence photon, photon rate changes, background photon, and background photon rate changes, are the most effective features. For example, open water is usually at sea level, wherein when the point moves from open water to sea ice or thin ice, the elevation level and height standard deviation, along with the other photon properties, also change. The point moves from along the IS2 track line path. Therefore, classifying a point in $n^{\text{th}}$ position depends on the information of the four surrounding point's $n-2$, $n-1$, $n+1$ and $n+2$ positions. Hence, we propose an LSTM \cite{hochreiter1997long} model for our classification task, which analyzes our data based on the progression data points. LSTM was originally developed to analyze sequential data where the output depends on the previous input. We use this property as the progression of data points, which simulates the change of sea ice cover (thick ice, thin ice, and open water.)
We employ a deep learning model implemented to classify geospatial data for our experiments.
The model begins with an LSTM layer, configured with 16 units and an Exponential Linear Unit (ELU) activation function, to effectively capture temporal dependencies within the input sequences, which are structured with a batch size of 5 and 6 features per time step. We include a dropout rate of 0.2 in the LSTM layer to maintain all information during the training phase. Following the LSTM, 7 Dense layers with 32, 96, 32, 16, 112, 48, and 64 units and ELU activation function are added to transform the learned features. The final layer is a Dense layer with three neurons, using a softmax activation function to output probabilities for the three classification categories. 
%

\subsubsection{MLP}
The MLP \cite{riedmiller2014multi} is a prevalent category of neural networks that are frequently employed in tasks that pertain to image and spatial data. To process this dataset, we need a deep learning model to get better classification results. 
We apply a fine-tuned MLP model to classify sea ice cover from the dataset. The model input is the same as the LSTM method. The model dense layer has 32 units, a RELU activation function, and a softmax activation function to output probabilities for the three classification categories in the final layer. 

Both the LSTM and MLP models are compiled with the Adam optimizer with a learning rate of 0.003 and focal loss as the loss function due to class imbalance of thick ice, thin ice and open water since we have more thick ice compared to the thin ice and open water regions in the Ross Sea.
Furthermore, we incorporate accuracy, F1 score, precision, and recall as performance metrics to ensure a comprehensive evaluation of the model’s classification capabilities. 



\subsubsection{Distributed Training using Horovod Framework}
Since the training dataset size is large and the deep learning models are computationally heavy, it takes some time to train the deep learning models. Hence, the application of distributed training techniques would help reduce the training time and scale the model training without losing accuracy. We train our models using the distributed training framework Horovod \cite{sergeev2018horovod}. We have employed \textit{synchronized data parallelism} to enhance the scalability of model training across multiple GPUs. It reduces the model training time and needs only a few number line modifications in the code. 
Instead of relying on one or multiple parameter servers, which are part of TensorFlow's built-in distribution strategy, we choose to leverage Horovod to aggregate and average gradients across multiple GPUs. The Horovod framework is a distributed deep-learning training framework that supports TensorFlow, Keras, PyTorch, and Apache MXNet. It allows us to distribute MLP and LSTM model training across multiple GPUs. To facilitate efficient inter-GPU communication, Horovod utilizes a ring-based all-reduce algorithm known for its bandwidth optimization and avoidance of system bottlenecks \cite{patarasuk2009bandwidth}. Coordination between processes in Horovod is achieved through MPI, with the Open-MPI-based wrapper being utilized for executing Horovod scripts.

To integrate our single-GPU implementation with the Horovod-based multiple-GPU distributed training, we follow the subsequent steps:
\begin{enumerate}
    \item Initialize Horovod using \textit{hvd.init()}.
    \item Assign a GPU to each of the TensorFlow processes.
    \item Wrap the TensorFlow optimizer with the Horovod optimizer using \textit{opt=hvd.DistributedOptimizer(opt)}. This Horovod optimizer handles gradient averaging using a ring-based all-reduce mechanism.
    \item Broadcast initial variable states from rank 0 to all other processes using\\ \textit{hvd.callbacks.BroadcastGlobalVariablesCallback(0)}.
\end{enumerate}
With the integration of Horovod, our model training accelerates significantly and becomes more scalable, all while maintaining accuracy.

\subsection{Deep Learning Model Inferencing}
High-precision LSTM and MLP models have been trained for the purpose of classifying sea ice using the IS2 ATL03 dataset.
\begin{figure}[htb]
        \centering
        \frame{\includegraphics[width=\linewidth]{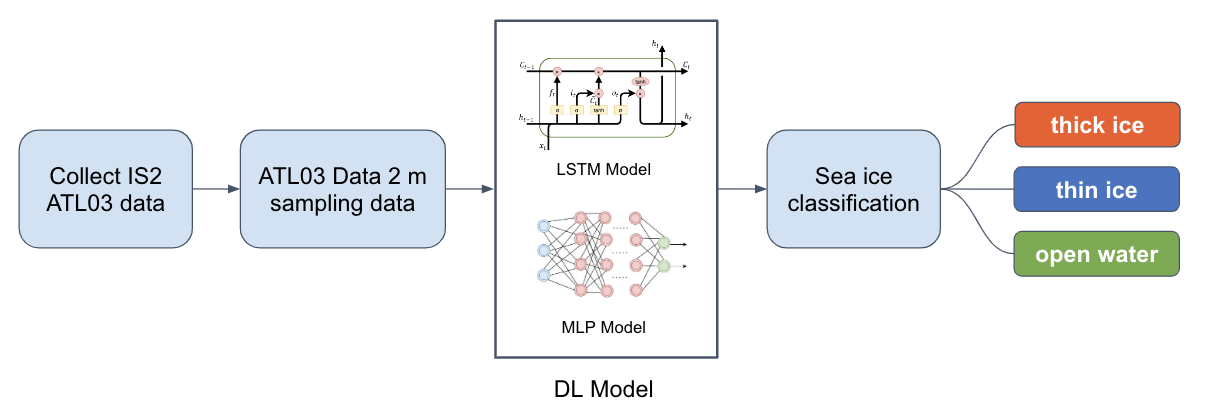}}
    \caption{Workflow for IS2 sea ice classification deep learning model inferencing.
    }
    \label{fig:is2_infer}
\end{figure}
In the inferencing workflow of the generic deep learning model, as depicted in Figure \ref{fig:is2_infer}, the initial step involves the acquisition of the original IS2 ATL03 data. Then, preprocess the data and 2m resampling data. Then, these processed and sampled data are utilized as input for the deep learning model during this inferencing phase. Finally, we obtain the sea ice types classified along ATL03 tracks as the output.


\subsection{Freeboard Computation}
The freeboard of sea ice is the distance (height difference) between the local sea level and the top of the sea ice (or snow if snow appears on sea ice). Freeboard is the basis for sea ice thickness calculations. 
After the sea ice classification, we calculate the sea ice freeboard using the classification results. 
First, we need to find the local sea surface height $h_{ref}$ of the region. Then, by subtracting the local sea surface height/elevation from the individual sea ice height/elevation $h_s$, we can find individual freeboard information, $h_f$, along the ATL03 tracks.

    \begin{equation}
        h_f = h_s - h_{ref}
    \end{equation}

\subsubsection{Local Sea Surface Detection}
As for local sea surface/level detection, we have chosen a sliding window-based strategy with a radius of 5km and a whole window size of 10km with a sliding overlap of 5km, which is similar to the way the original IS2 ATL10 retrieving local sea surface \cite{kwok2022icesat}.
We use open water or lead region for a particular window of a 5 km radius (10 km length) to select the local sea level for that window region. 
However, if there is no open water for a particular window, we do a linear interpolation with respect to the nearest local sea surface to derive the local sea surface for that area.

For calculating the local sea surface, we have tested four different approaches,
\begin{enumerate}
    \item Minimum Elevation: Finding the minimum elevation of open water from a given window of a 5 km radius (10 km length).
    \item Average Elevation: Finding the average elevation of open water from a given window of a 5 km radius (10 km length).
    \item Nearest Minimum Elevation: Finding the nearest minimum elevation of open water from a given window within a 5 km radius (10 km length).
    \item The sea surface equation formulated by NASA involves the utilization of standard deviation and various other elements to determine the elevation of the sea surface within a 10km length segment of the data track. 
    For each of these segments, we first calculate the mean lead height $\hat{h_{lead\_w}}$ and estimated error $\hat{\sigma^2_{lead\_w}}$ of a single open water lead from candidate points as follows:

    \begin{equation}
        \hat{h}_{lead\_w} = \sum_{i=1}^{N_s} a_i h_i
        \text{ and }
        \hat{\sigma^2}_{lead\_w} = \sum_{i=1}^{N_s} \alpha_i^2 \sigma_i^2    
    \end{equation}

    where, $\alpha_i = \frac{w_i}{\sum_{i=1}^{N_s} w_i}$ and $w_i = \exp \left( - \frac{h_i - h_{min}}{\sigma_i} \right)^2 $.

    In this equation, $h_i$ is the surface height, and $\sigma_i^2$ is the error variance of each 2m sampled height estimate from open water candidate points, respectively. 
    Here, $h_{min}$ stands for the minimum height of each lead group point, and $N_s$ denotes the number of samples forming the individual open-water lead.
 
    After these open water leads are identified, the sea reference height $\hat{h}_{ref}$ is calculated following NASA's \cite{kwok2022icesat} equations that are done based on all the open water lead segments in the 10km segment: 
    
    \begin{equation}
        \hat{h}_{ref} = \sum_{i=1}^{N_l} a_i \hat{h}_{lead\_w(i)} 
        \text{ and }
        \hat{\sigma^2}_{ref} = \sum_{i=1}^{N_l} \alpha_i^2 \sigma_{lead\_w(i)}^2
    \end{equation}

    where, $\alpha_i = \frac{\frac{1}{\sigma_{lead\_w(i)}^2}}{\sum_{j=1}^{N_l}\frac{1}{\sigma_{lead\_w(i)}^2}}.$ Here, $N_l$ denotes the number of leads in a 10km segment.

\end{enumerate}

We compare our results of local sea level with the ATL07 and ATL10 to decide the best one for this study.












\subsubsection{Scaled Freeboard Computation}
To accelerate the freeboard computation, we use the map-reduce-based PySpark framework similar to our PySpark-based auto-labeling of IS2. 
Since we have a large volume of data, partitioning it and distributing the computation across multiple machines would improve scalability and processing speed. By dividing the data into smaller chunks and handling these chunks in parallel on different machines, we can process the data more efficiently and reduce overall computation time. 
Spark facilitates the partitioning of large datasets and the distribution of workloads across multiple worker nodes. This capability allows us to implement data parallelism, thereby improving the efficiency and scalability of our project.


%% file: sections/experiments.tex



\subsection{Experimental Setup and Evaluation Metrics}


We utilize the Google Cloud Dataproc (GCD) service on the Google Cloud Platform for PySpark-based experiments. There, we use a cluster of four nodes with one master node and three worker nodes. Each of the Intel N2 Cascade Lake computers is equipped with four cores.

As for the DL models, first, we divide the dataset into 80\% training dataset and 20\% test dataset. Then, we organize the data into batches for the MLP and LSTM models. We use the Adam optimizer, dropouts of 0, 0.2, and 0.3 in different convolutional layers, and epochs of 20, 30, and 40 to observe the changes. Both our LSTM and MLP model has a dropout rate of 0.2, a batch size of 32, and the number of epochs is 20. 
Detailed results are reported in the following parts. 

Since the model training is computationally heavy, we have applied Horovod-based distributed training. We have utilized an NVIDIA DGX A100 machine with dual CPUs, each with four A100 GPUs.

To validate the results of our algorithm, 
    we compute the accuracy, precision, recall, and F1 score to obtain a comprehensive and balanced evaluation of the model's performance.
    For the two models, MLP and LSTM, we evaluate the models using a 20\% validation dataset to find the overall classification accuracy of these two models.  
    
    

\subsection{IS2 Auto-labeling} 

\subsubsection{IS2 Auto-labeling Speedup}
The GCD service is utilized alongside the PySpark framework for thin cloud and shadow-filtered autolabeling of IS2 data. 
This technique was applied for the annotation of S2 data, which was subsequently used for training deep learning models \cite{iqrah2023toward}, \cite{iqrah2024parallel}. Then, based on the overlapped IS2 and S2 Data, IS2 data is auto-labeled. 
IS2 auto-labeling speedup results for the PySpark-based approach are as follows, in table \ref{tab:auto_label_spark}.
\begin{table}[htb]
\centering
\caption{PySpark-based IS2 auto-labeling scalability over Google Cloud.}
\label{tab:auto_label_spark}
\begin{tabular}{ |>{\centering\arraybackslash}p{0.115\linewidth}||>{\centering\arraybackslash}p{0.06\linewidth}||>{\centering\arraybackslash}p{0.06\linewidth}||>{\centering\arraybackslash}p{0.06\linewidth}||>{\centering\arraybackslash}p{0.08\linewidth}||>{\centering\arraybackslash}p{0.11\linewidth}||>{\centering\arraybackslash}p{0.11\linewidth}|}
\hline
\textbf{Executors} & \textbf{Cores} & \textbf{Load Time (s)} & \textbf{Map Time (s)} & \textbf{Reduce Time (s)} & \textbf{Speed-up Load} & \textbf{Speed-up Reduce} \\ \hline
1                  & 1              & 108                & 0.4               & 390                  & 1                      & 1                        \\ \hline
1                  & 2              & 58                 & 0.4               & 174                  & 1.86                   & 2.24                     \\ \hline
1                  & 4              & 33                 & 0.3               & 72                   & 3.27                   & 5.42                     \\ \hline
2                  & 1              & 56                 & 0.3               & 156                  & 1.93                   & 2.5                      \\ \hline
2                  & 2              & 31                 & 0.3               & 84                   & 3.48                   & 4.64                     \\ \hline
2                  & 4              & 19                 & 0.3               & 41                   & 5.68                   & 9.51                     \\ \hline
4                  & 1              & 31                 & 0.2               & 78                   & 3.48                   & 5                        \\ \hline
4                  & 2              & 17                 & 0.2               & 39                   & 6.35                   & 10                       \\ \hline
4                  & 4              & 12                 & 0.3               & 24                   & \textbf{9}                      & \textbf{16.25}                    \\ \hline
\end{tabular}
\end{table}
A speedup of up to 16.25 times is attained in the execution of this workflow, as illustrated in Table \ref{tab:auto_label_spark}. Additionally, there is a significant improvement in data loading speed when utilizing numerous machines, with a maximum speedup of up to 9 times. 
The auto-labeling of IS2 data is highly scalable due to independent data point processing, albeit fine-grained. This is easily parallelized using PySpark (map-reduce framework). PySpark is utilized here to parallelize and scale the auto-labeling of the IS2 data on different architectures. Along with a single multi-core machine, it is scaled over multiple heterogeneous machines in a GCD cluster with 9.0x data loading and 16.25x map-reduce processing speedup. Since the PySpark-based approach supports larger clusters for distributing the auto-labeling procedure, it points to a potential for scaling over much larger data in the future.

\subsection{Model Training Results}

\subsubsection{Model Accuracy}

\begin{table}[htb]
\centering
\caption{DL models sea ice classification accuracy over IS2 ATL03 Antarctic summer datasets. }
\label{tab:dl_accuracy}
\begin{tabular}{ |>{\centering\arraybackslash}p{0.18\linewidth}||>{\centering\arraybackslash}p{0.14\linewidth}||>{\centering\arraybackslash}p{0.14\linewidth}||>{\centering\arraybackslash}p{0.12\linewidth}||>{\centering\arraybackslash}p{0.12\linewidth}|}
\hline
\textbf{Model} & \textbf{Accuracy} & \textbf{Precision} & \textbf{Recall} & \textbf{F1 score} \\ \hline
\textbf{MLP}   & 91.80             & 91.80              & 91.80           & 91.79             \\ \hline
\textbf{LSTM}  & 96.56             & 97.00              & 96.09           & 96.54             \\ \hline
\end{tabular}
\end{table}

\begin{figure}
    \centering
    \includegraphics[width=0.5\linewidth]{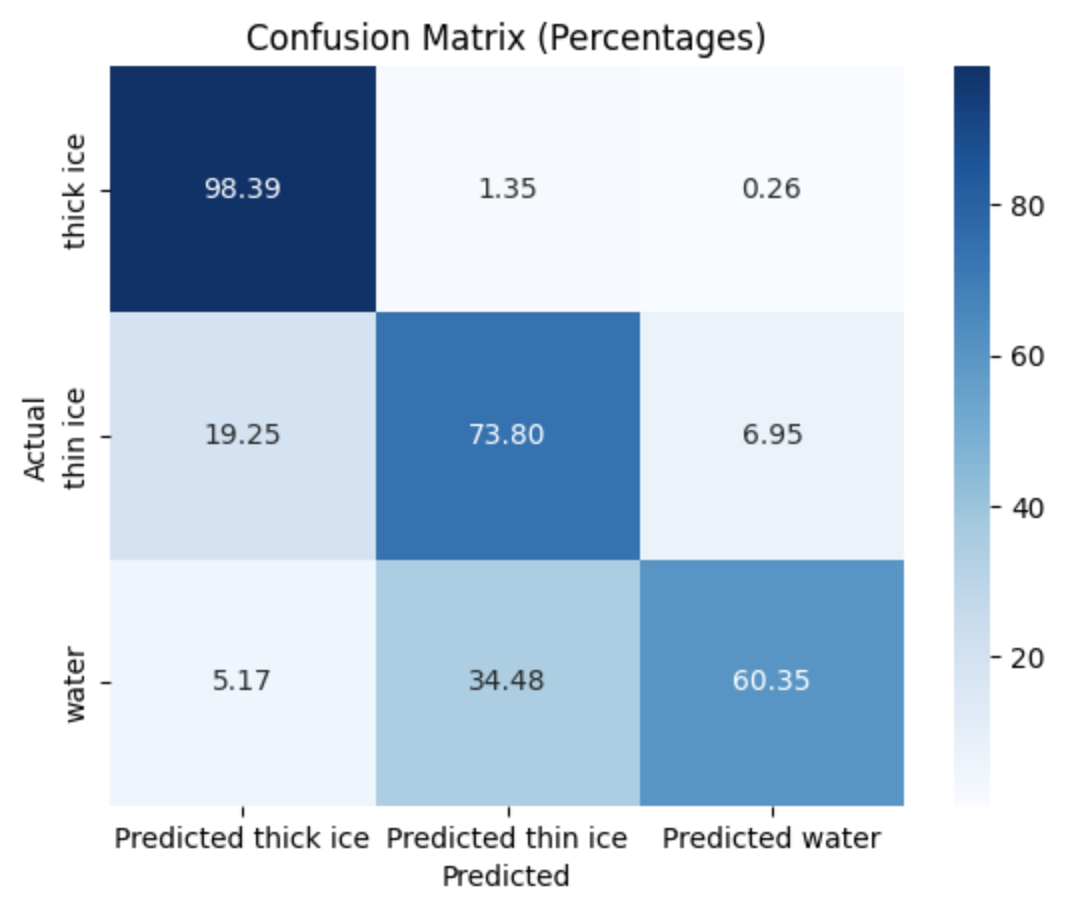}
    \caption{Sea-ice Classification Confusion Matrix}
    \label{fig:confusion_mat}
\end{figure}

The accuracy comparison of the LSTM and MLP model is presented in Table \ref{tab:dl_accuracy} for IS2 ATL03 data. Basically, the two methods show over 90\% accuracy results. However, the LSTM has a better accuracy of 96.56\% compared to the MLP with 91.80\%. Therefore, the LSTM is used for sea ice classification in the study. Moreover, the figure \ref{fig:confusion_mat} represents a confusion matrix consisting of detailed individual classification accuracy of thick ice, thin ice, and open water with 98.39\%, 73.80\%, and 60.25\%, respectively.

\begin{table}[htb]
\caption{Distributed DL model training using Horovod framework on DGX A100 cluster.}
\label{tab:dl_training_horovod}
\begin{tabular}{ |>{\centering\arraybackslash}p{0.10\linewidth}||>{\centering\arraybackslash}p{0.12\linewidth}||>{\centering\arraybackslash}p{0.25\linewidth}||>{\centering\arraybackslash}p{0.12\linewidth}||>{\centering\arraybackslash}p{0.12\linewidth}|}
\hline
\textbf{No. of GPUs} & \textbf{Time (s)} & \textbf{Time (s)/Epoch} & \textbf{Data/s} & \textbf{Speedup}     \\ \hline
1          & 280.72                        & 5.5                                 & 585.88                       & 1.00    \\ \hline
2          & 143.22                        & 2.778                               & 1160.81                      & 1.96    \\ \hline
4          & 73.68                         & 1.45                                & 2229.56                      & 3.81    \\ \hline
6          & 49.42                         & 0.97                                & 3330.03                      & 5.68    \\ \hline
8          & 38.72                         & 0.79                                & 4248.56                      & 7.25   \\ \hline
\end{tabular}
\end{table}

\begin{figure}[htb]
        \centering
        \begin{subfigure}[b]{0.24\textwidth}
            \centering
            \frame{\includegraphics[width=\textwidth]{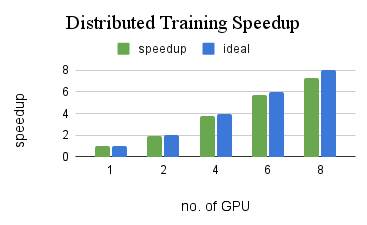}}
            \caption{}
            \label{fig:distributed_training_speedup}
        \end{subfigure}
        \begin{subfigure}[b]{0.24\textwidth}
            \centering
            \frame{\includegraphics[width=\textwidth]{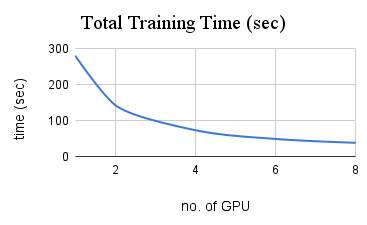}}
            \caption{}
            \label{fig:total_train_time}
        \end{subfigure}
        \par\medskip
        \begin{subfigure}[b]{0.24\textwidth}
            \centering
            \frame{\includegraphics[width=\textwidth]{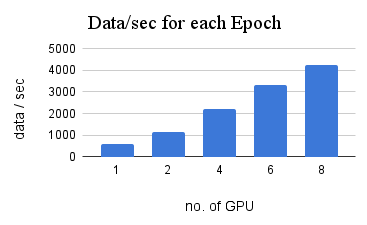}}
            \caption{}
            \label{fig:data_sec_per_epoch}
        \end{subfigure}
        \begin{subfigure}[b]{0.24\textwidth}
            \centering
            \frame{\includegraphics[width=\textwidth]{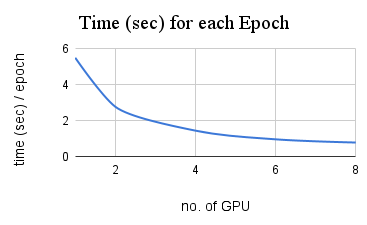}}
            \caption{}
            \label{fig:time_per_epoch}
        \end{subfigure}
    \caption{Distributed model training via Horovod framework, 
    (a) distributed training speedup, 
    (b) total training time over multiple GPUs, 
    (c) data processed per second for each epoch and 
    (d) time for each epoch.}
    \label{fig:horovod_training_chart}
\end{figure}

\subsubsection{Distributed Model Training Speedup}
Table \ref{tab:dl_training_horovod} shows the scaled and distributed model training results. We train our LSTM model in the DGX A100 cluster using the Horovod framework. We calculate the time for our Horovod-based model training in 1, 2, 4, 6, and 8 GPU setups and a batch size of 32. The training time is reduced from 280.72s for 1 GPU to 38.72s for 8 GPU, gaining up to 7.25x speedup. We have trained our model for 20 epochs, and for each epoch, we achieve up to 4248.56 image data/s throughput within 0.79s on 8 GPUs compared to 585.88 image data/s throughput with 5.5s on a single GPU. Figure \ref{fig:horovod_training_chart} shows the performance results of distributed model training via Horovod over multiple GPUs. 
Here, we observe that the distributed training speedup and the throughput growth rate are almost close to linear, which is ideal with the increased number of GPUs.
On the other hand, the total training time and the time per epoch decreasing rate are high initially; however, eventually, they slow down with the increased number of GPUs. 
During training, the bottleneck arises from data preprocessing and subsequent batch preparation, resulting in GPU starvation. As a result, we are not achieving optimal speedup or throughput performance.

%

\subsubsection{Sea-ice Classification Comparison}
\begin{figure}[ht]
        \centering
        \begin{subfigure}[b]{\linewidth}
            \centering
            \includegraphics[width=\textwidth]{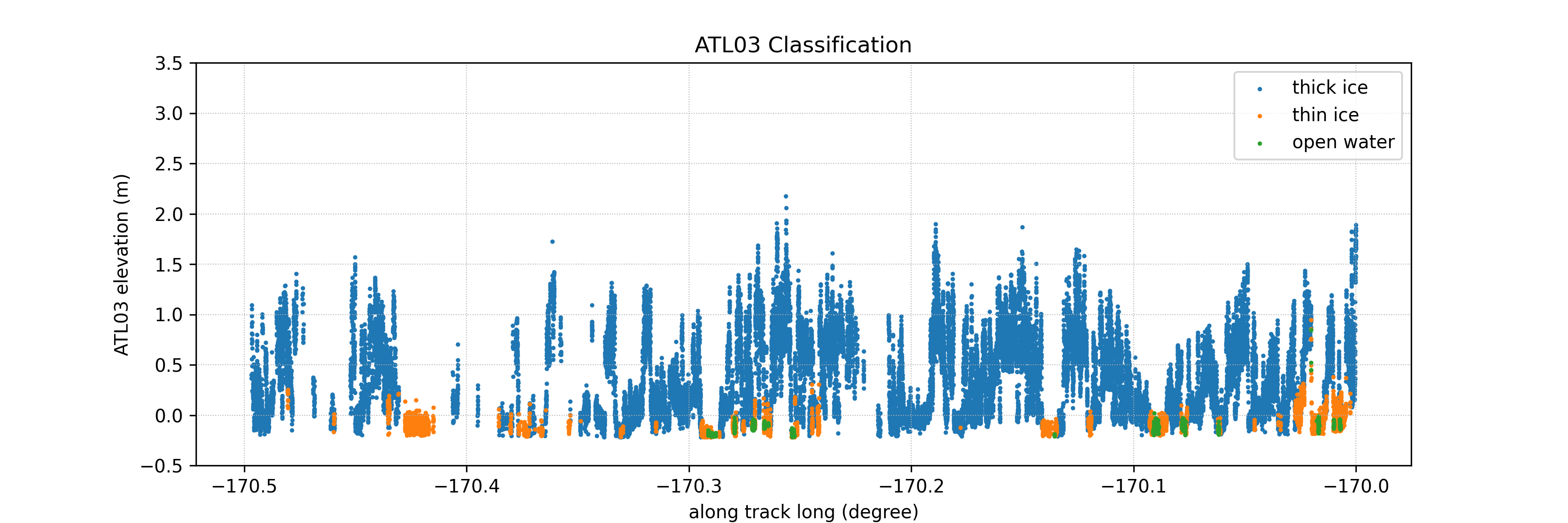}
            \caption{Sea ice classification on ATL03}
            \label{fig:elev04_atl03}
        \end{subfigure}

        \begin{subfigure}[b]{\linewidth}
            \centering
            \includegraphics[width=\textwidth]{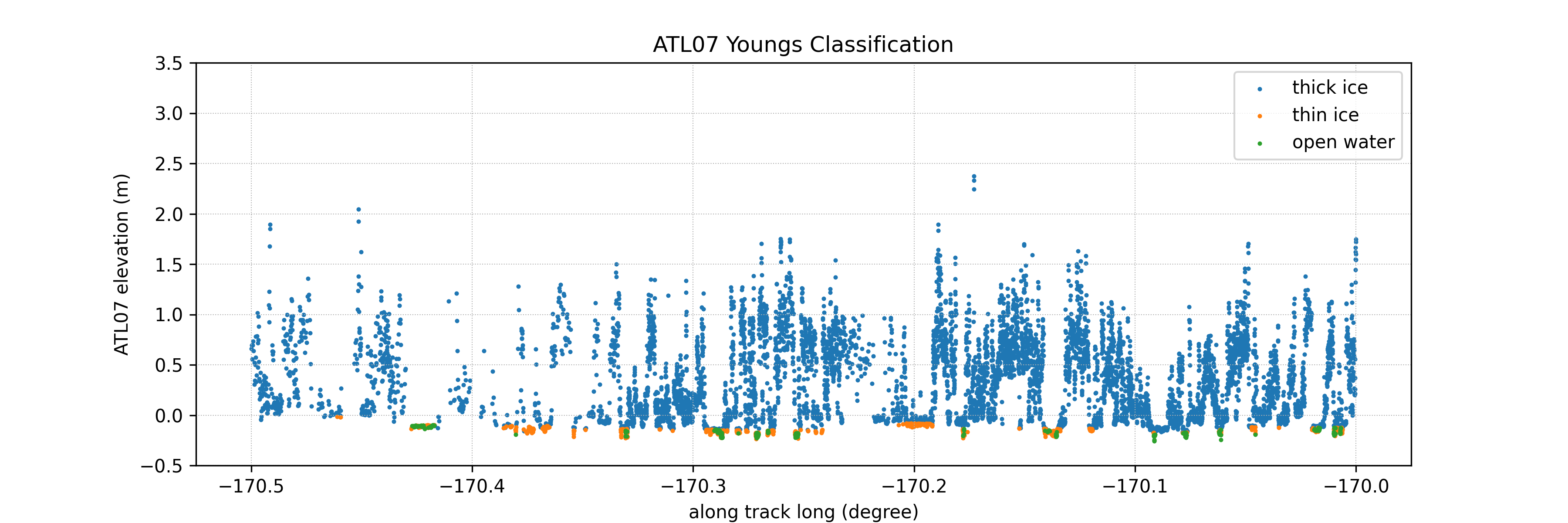}
            \caption{Sea ice classification on ATL07 Koo\_method \cite{koo2023sea}}
            \label{fig:elev_atl07_1}
        \end{subfigure}
    \caption{Sea ice classification comparison of ATL03 and ATL07 (Koo\_method \cite{koo2023sea}) of IS2 track \textit{20191104195311\_05940510\_gt2r}. Here, thick ice is blue, thin ice is green, and open water is orange}
    \label{fig:classification_atl03_atl07_1}
\end{figure}

\begin{figure}[ht]
        \centering
        \begin{subfigure}[b]{\linewidth}
            \centering
            \includegraphics[width=\textwidth]{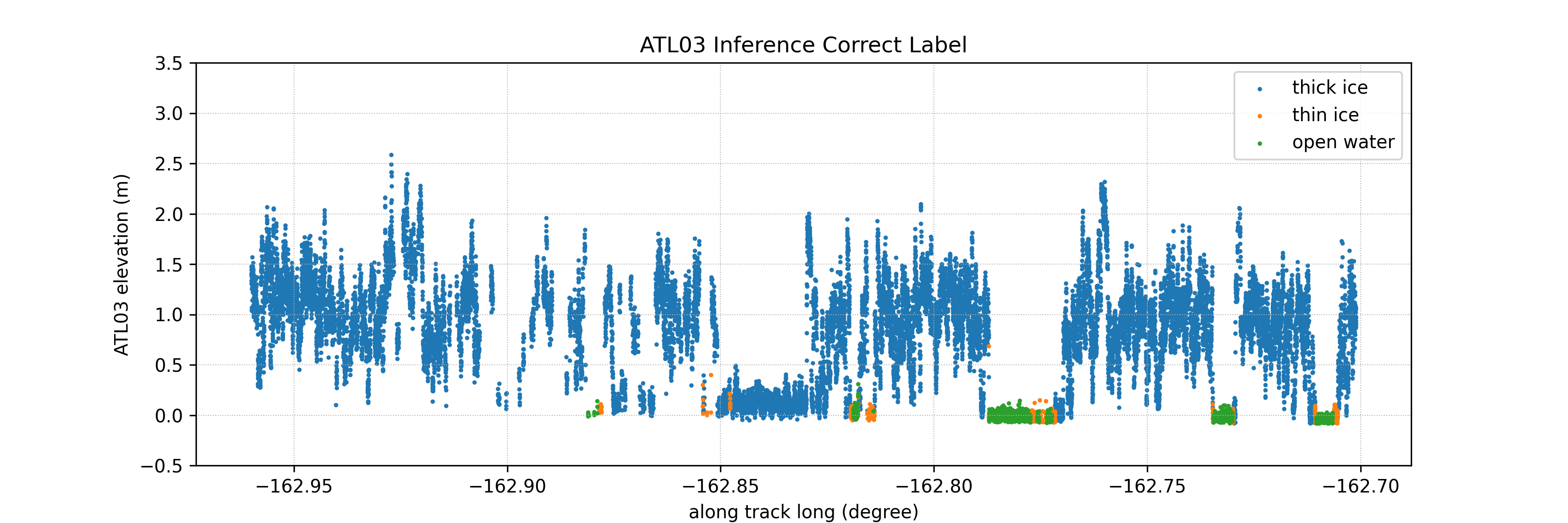}
            \caption{Sea ice classification on ATL03}
            \label{fig:elev26_atl03}
        \end{subfigure}

        \begin{subfigure}[b]{\linewidth}
            \centering
            \includegraphics[width=\textwidth]{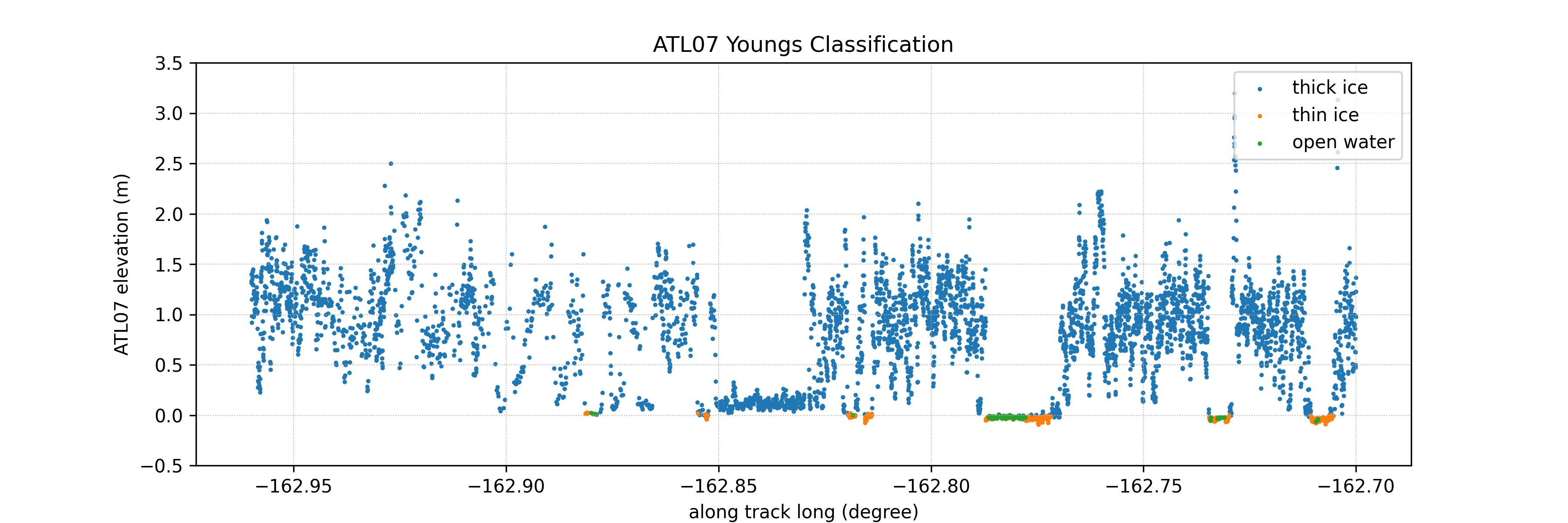}
            \caption{Sea ice classification on ATL07 Koo\_method \cite{koo2023sea}}
            \label{fig:elev_atl07}
        \end{subfigure}
    \caption{Sea ice classification comparison of ATL03 and ATL07 (Koo\_method \cite{koo2023sea}) of IS2 track \textit{20191126182014\_09290510\_gt2r}. Here, thick ice is blue, thin ice is green, and open water is orange}
    \label{fig:classification_atl03_atl07_2}
\end{figure}

After the LSTM-based sea ice classification of ATL03 data, different types of sea ice (thick ice, thin ice and open water) classification over IS2 ATL03 product are compared with IS2 ALT07's sea ice classification product. We did the ATL03 classification using the LSTM model, where \cite{koo2023sea} used the MLP model for the ATL07 classification. ATL03 and ATL07 data product's sea ice classification is plotted in figure \ref{fig:classification_atl03_atl07_1} and \ref{fig:classification_atl03_atl07_2}. 
%
Based on the results in figure \ref{fig:classification_atl03_atl07_1} and \ref{fig:classification_atl03_atl07_2}, we can see that our ATL03 data product's sea ice classification in figure \ref{fig:elev26_atl03} and \ref{fig:elev04_atl03} are more dense than ATL07. It provides a higher resolution sea ice classification product than the ATL07 sea ice classification in figures \ref{fig:elev_atl07} and \ref{fig:elev_atl07_1}.

\subsection{Freeboard Computation}
For freeboard calculation, the first step is to find the local sea surface and then calculate the freeboard based on the height difference of sea ice with the derived local sea surface. 
IS2 ATL03 freeboard computation results, along with the PySpark-based approach, are as follows,

\subsubsection{Local Sea Level Comparison}
\begin{figure}[ht]
        \centering

        \begin{subfigure}[b]{\linewidth}
            \centering
            \includegraphics[width=\textwidth]{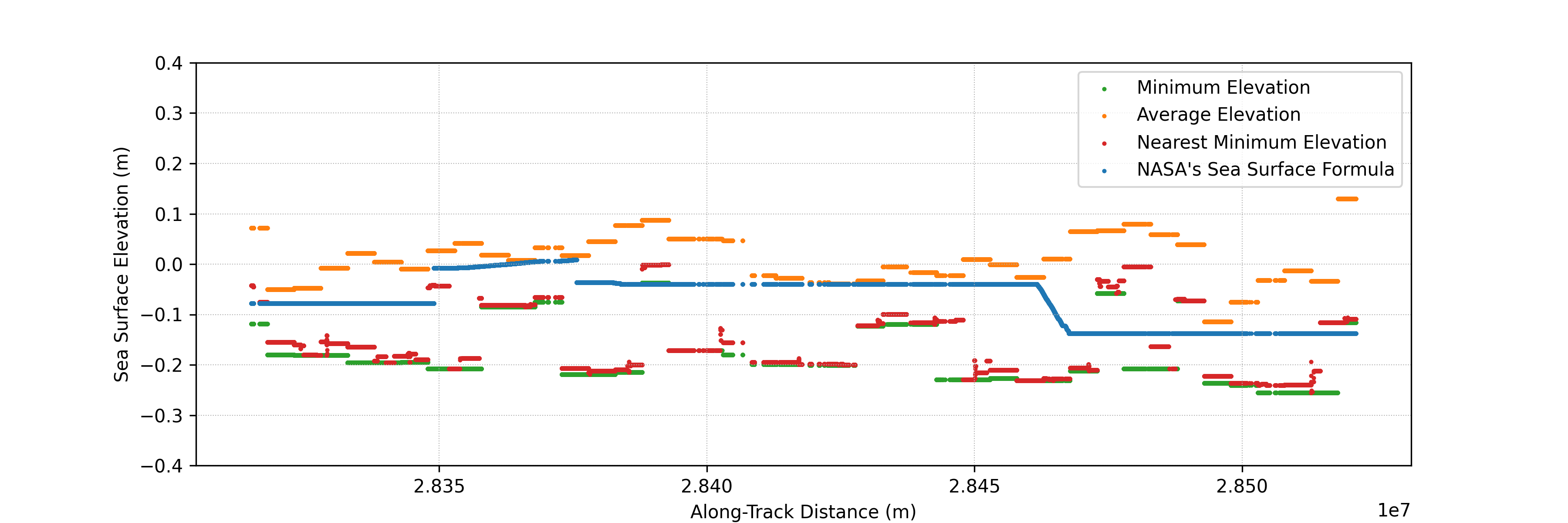}
            \caption{Local sea surface using four different methods from ATL03}
            \label{fig:ss04_1}
        \end{subfigure}

        \begin{subfigure}[b]{\linewidth}
            \centering
            \includegraphics[width=\textwidth]{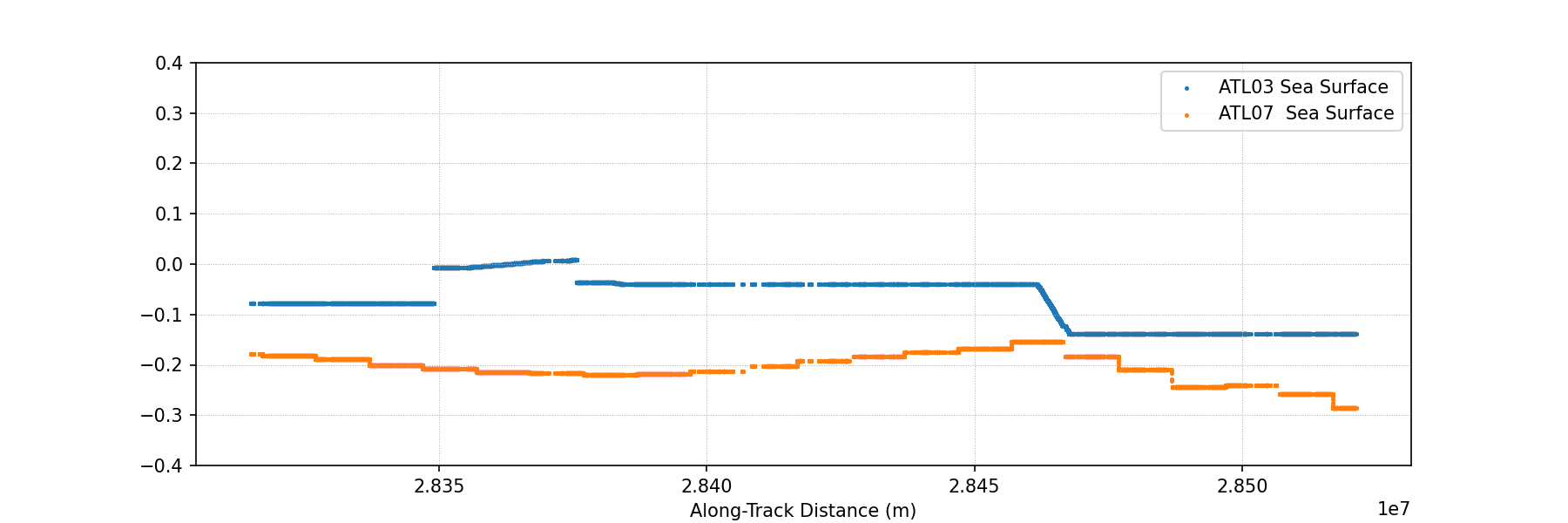}
            \caption{Local sea surface of ATL03 and ATL07 (Koo\_method \cite{koo2023sea}}
            \label{fig:ss04_2}
        \end{subfigure}
    \caption{Local sea surface detection based on four different methods from ATL03 (a) and comparison based on ATL03 (this paper) and ATL07 (Koo\_method \cite{koo2023sea}) (b) over IS2 track \textit{20191104195311\_05940510\_gt2r}.} 
    \label{fig:localSS_1}
\end{figure}

\begin{figure}[ht]
        \centering
        \begin{subfigure}[b]{\linewidth}
            \centering
            \includegraphics[width=\textwidth]{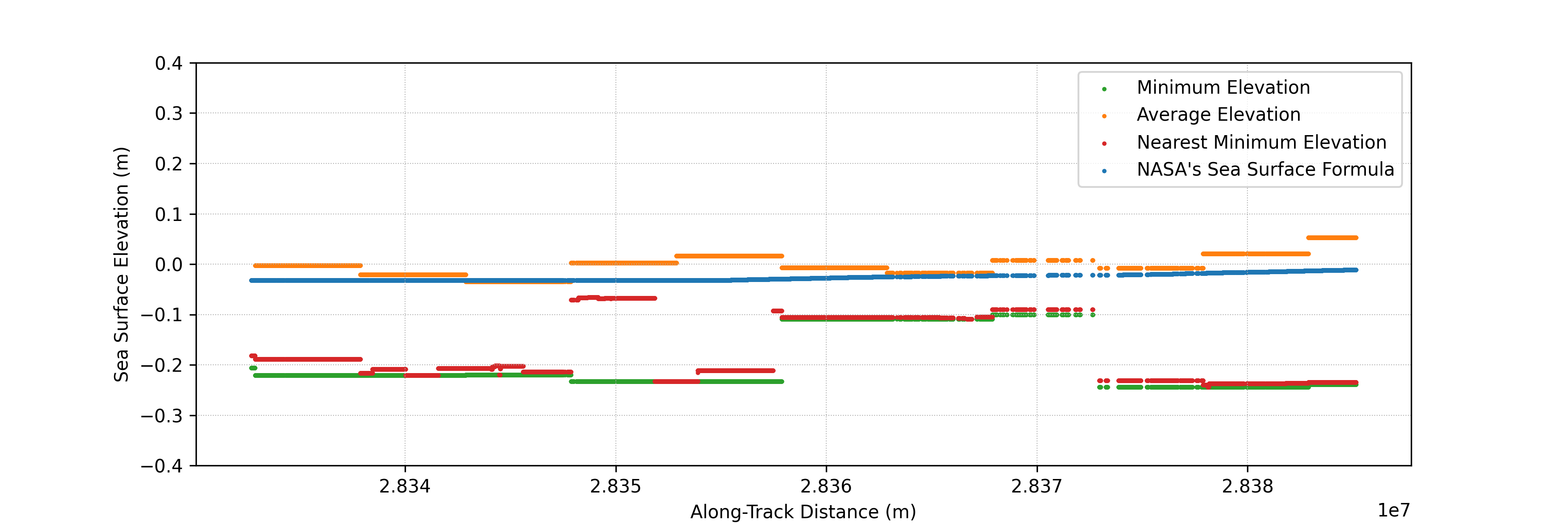}
            \caption{Local sea surface using four different methods from ATL03}
            \label{fig:ss26_1}
        \end{subfigure}

        \begin{subfigure}[b]{\linewidth}
            \centering
            \includegraphics[width=\textwidth]{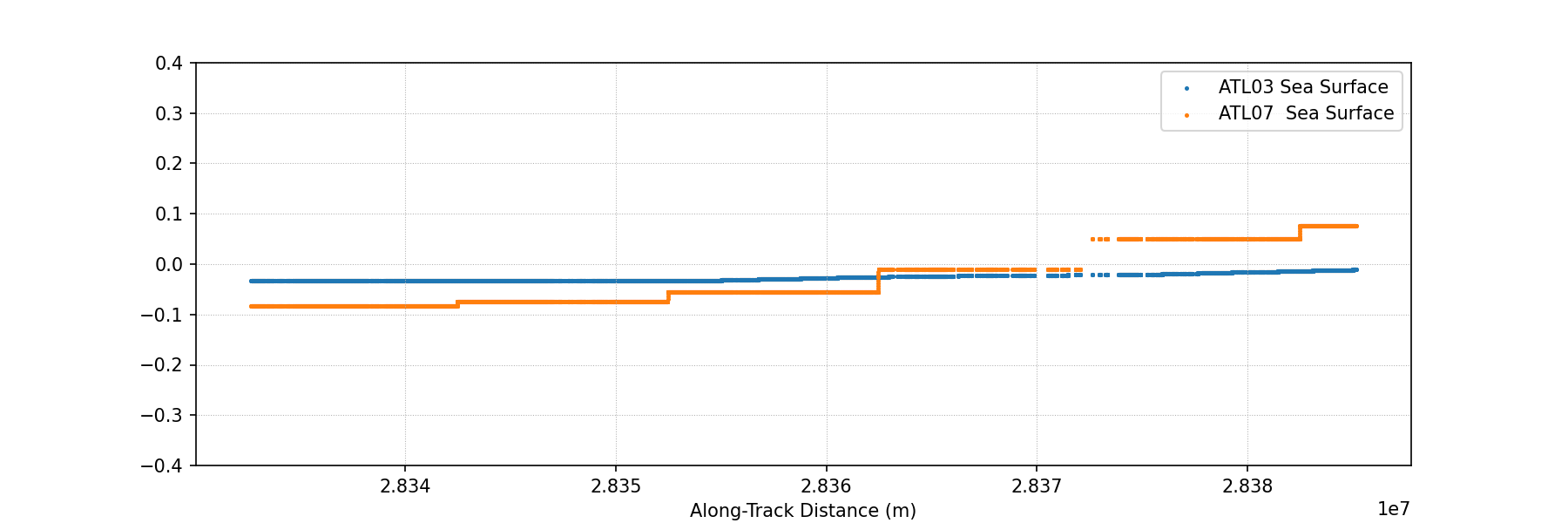}
            \caption{Local sea surface of ATL03 and ATL07 (Koo\_method \cite{koo2023sea}}
            \label{fig:ss26_2}
        \end{subfigure}
    \caption{Local sea surface detection based on four different methods from ATL03 (a) and comparison based on ATL03 (this paper) and ATL07 (Koo\_method \cite{koo2023sea}) (b) over IS2 track \textit{20191126182014\_09290510\_gt2r}.} 
    \label{fig:localSS_2}
\end{figure}

We have applied four different techniques for local sea surface detection: i) Minimum Elevation, ii) Average Elevation, iii) Nearest Minimum Elevation, and iv) The sea surface equation formulated by NASA.
To compare these different local sea surface detection methods, we plot the four different types of local sea surface products shown in figure \ref{fig:localSS_1} and \ref{fig:localSS_2}.
In Figure \ref{fig:ss04_1} and \ref{fig:ss26_1}, we observe that the sea surface detection using ATL03 data, based on NASA's sea surface detection formula (represented by the blue line), is a better technique compared to other methods, as it provides a smoother local sea surface. As a result, we select this sea surface based on NASA's sea surface detection formula one as our primary local sea surface. We also compare this sea surface with Koo\_method \cite{koo2023sea}, ATL07 product and saw that these two have a similar sea surface, and the difference between them is little over 0.1m. This comparison is illustrated in figure \ref{fig:ss04_2} and \ref{fig:ss26_2}. 

\begin{figure}[!htb]
        \centering
        \begin{subfigure}[b]{\linewidth}
            \centering
            \includegraphics[width=\textwidth]{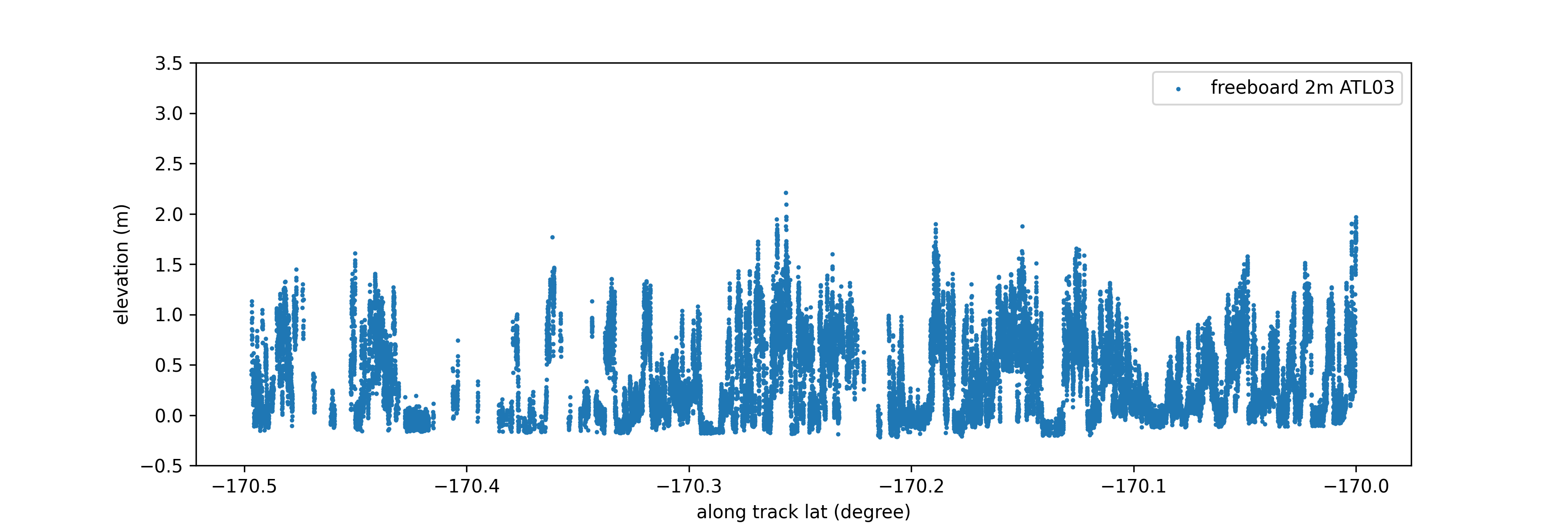}
            \caption{Freeboard from ATL03}
            \label{fig:freeboard_03_1}
        \end{subfigure}
        \begin{subfigure}[b]{\linewidth}
            \centering
            \includegraphics[width=\textwidth]{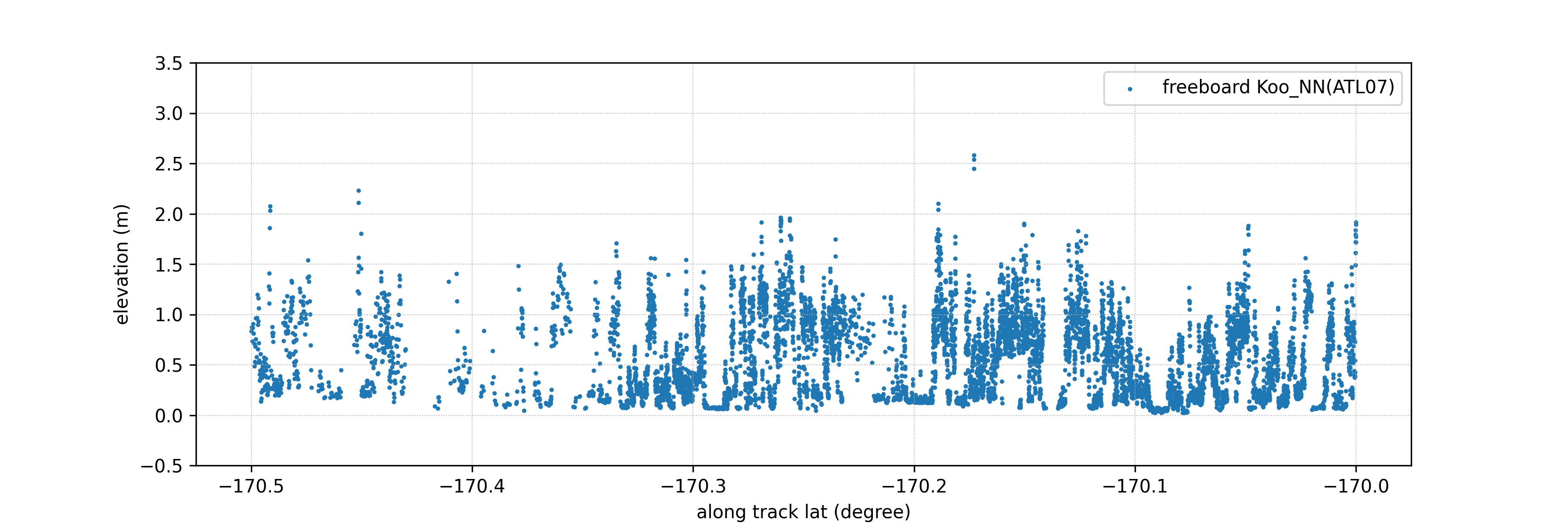}
            \caption{Freeboard from ATL07 (Koo\_method \cite{koo2023sea})}
            \label{fig:freeboard_07_1}
        \end{subfigure}
        \begin{subfigure}[b]{0.9\linewidth}
            \centering
            \includegraphics[width=\textwidth]{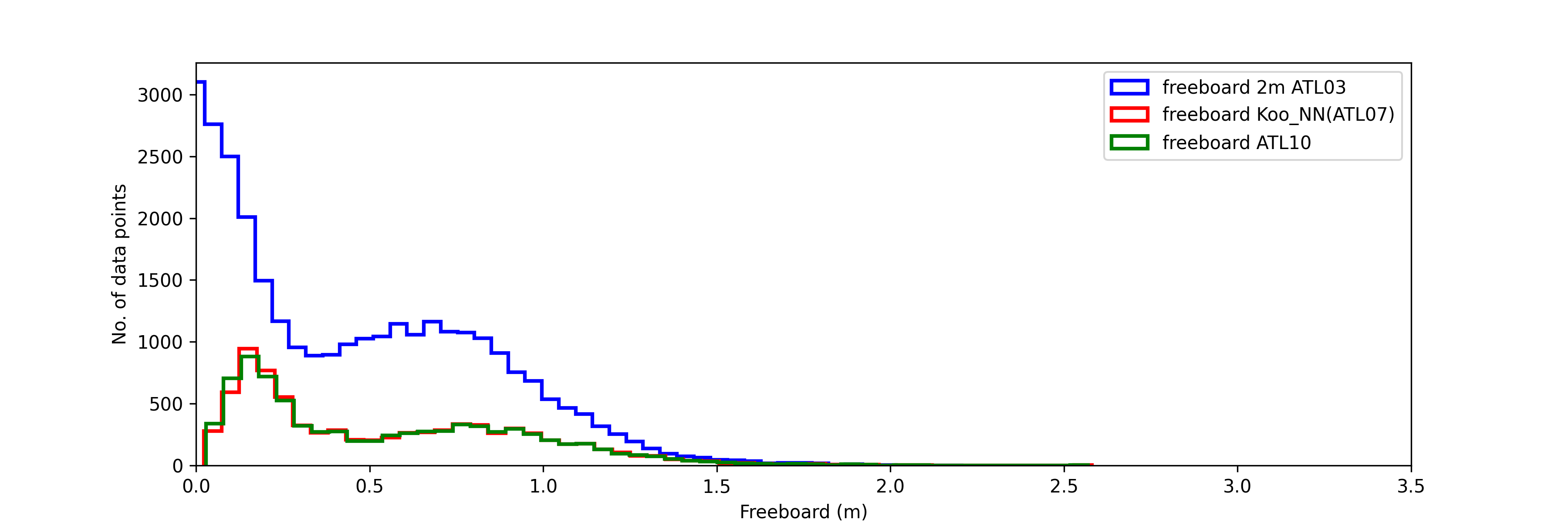}
            \caption{Freeboard distributions from ATL03, ATL07 (Koo\_method), and ATL10}
            \label{fig:freeboard_hist04_atl_03_07_10}
        \end{subfigure}
        \begin{subfigure}[b]{\linewidth}
            \centering
            \includegraphics[width=\textwidth]{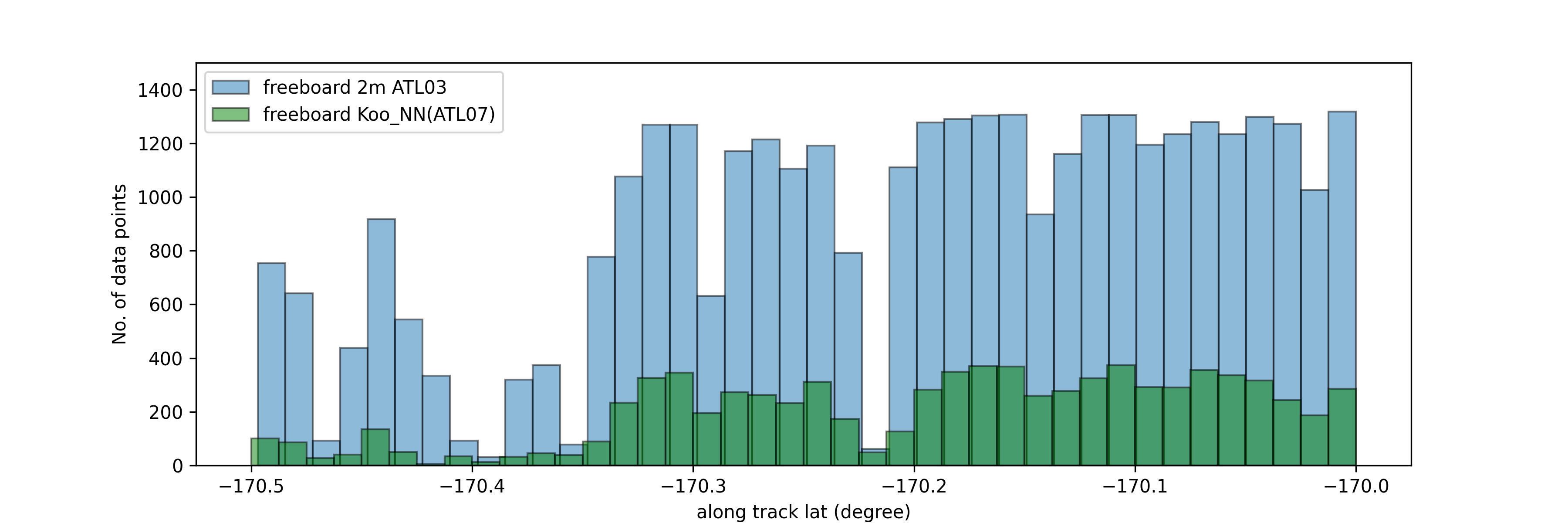}
            \caption{Point density difference between ATL03 and ATL07 (Koo\_method)}
            \label{fig:hist04_atl_03_10}
        \end{subfigure}
    \caption{Freeboard from this study (a) and from ATL07 (Koo\_method) (b), freeboard distributions (c) from this study, ATL07 (Koo\_method) and ATL10, 0, and (d) point density between this study and ATL07 (Koo\_method) along the IS2 track \textit{20191104195311\_05940510\_gt2r}}
    \label{fig:freeboard_comp1}
\end{figure}

\begin{figure}[!htb]
        \centering
        \begin{subfigure}[b]{\linewidth}
            \centering
            \includegraphics[width=\textwidth]{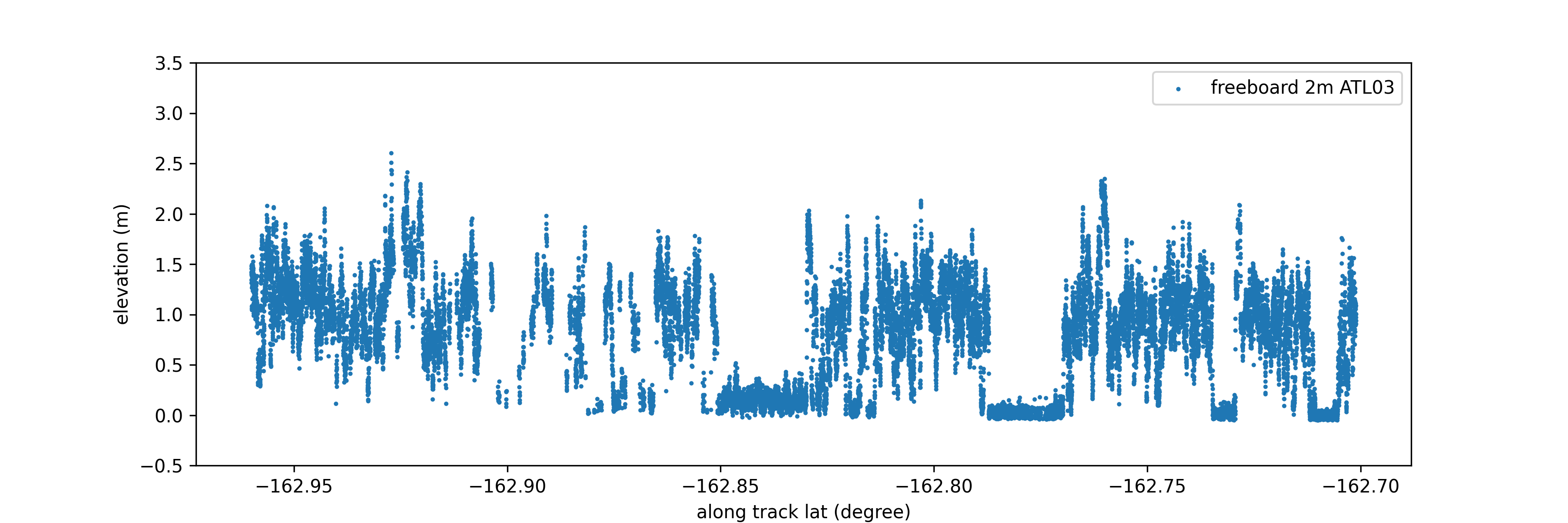}
            \caption{Freeboard from ATL03}
            \label{fig:freeboard_03}
        \end{subfigure}
        \begin{subfigure}[b]{\linewidth}
            \centering
            \includegraphics[width=\textwidth]{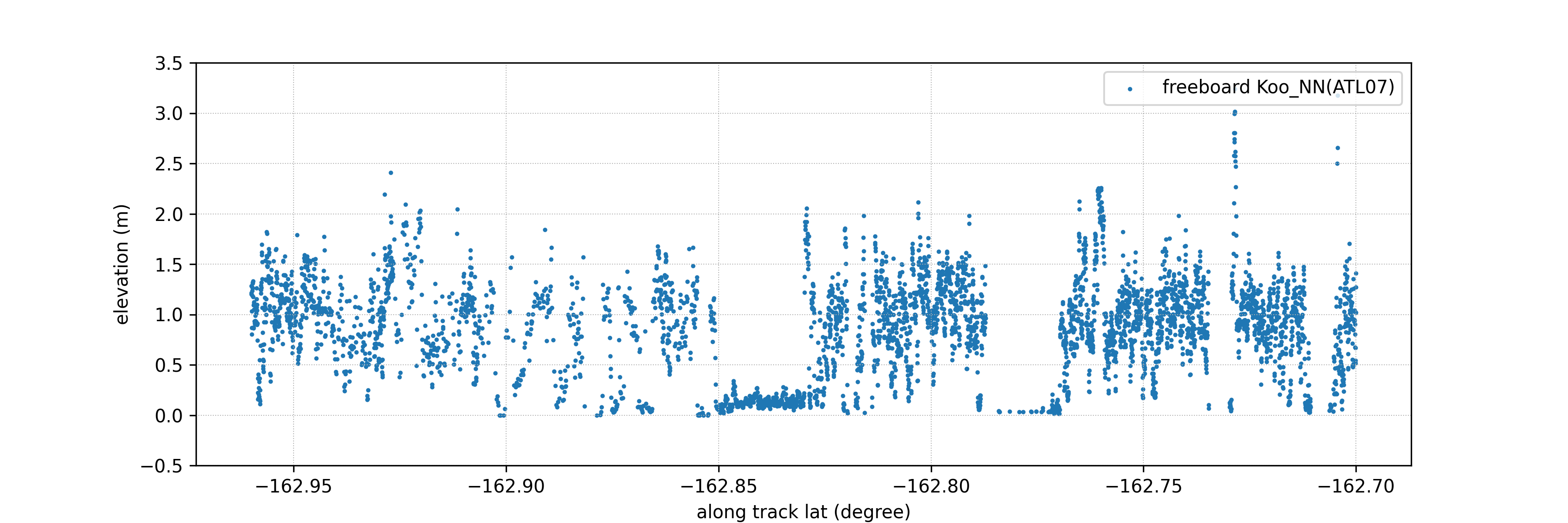}
            \caption{Freeboard from ATL07 (Koo\_method \cite{koo2023sea})}
            \label{fig:freeboard_07}
        \end{subfigure}
        \begin{subfigure}[b]{0.9\linewidth}
            \centering
            \includegraphics[width=\textwidth]{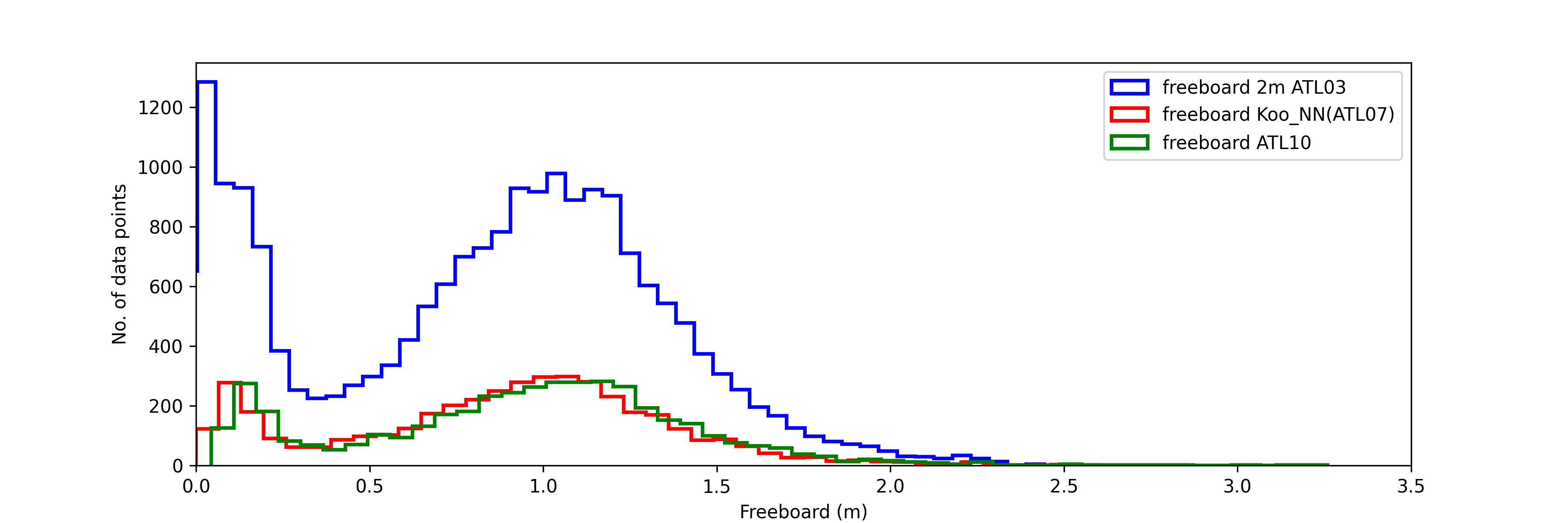}
            \caption{Freeboard distributions from ATL03, ATL07 (Koo\_method), and ATL10}
            \label{fig:freeboard_hist26_atl_03_07_10_v1}
        \end{subfigure}
        \begin{subfigure}[b]{\linewidth}
            \centering
            \includegraphics[width=\textwidth]{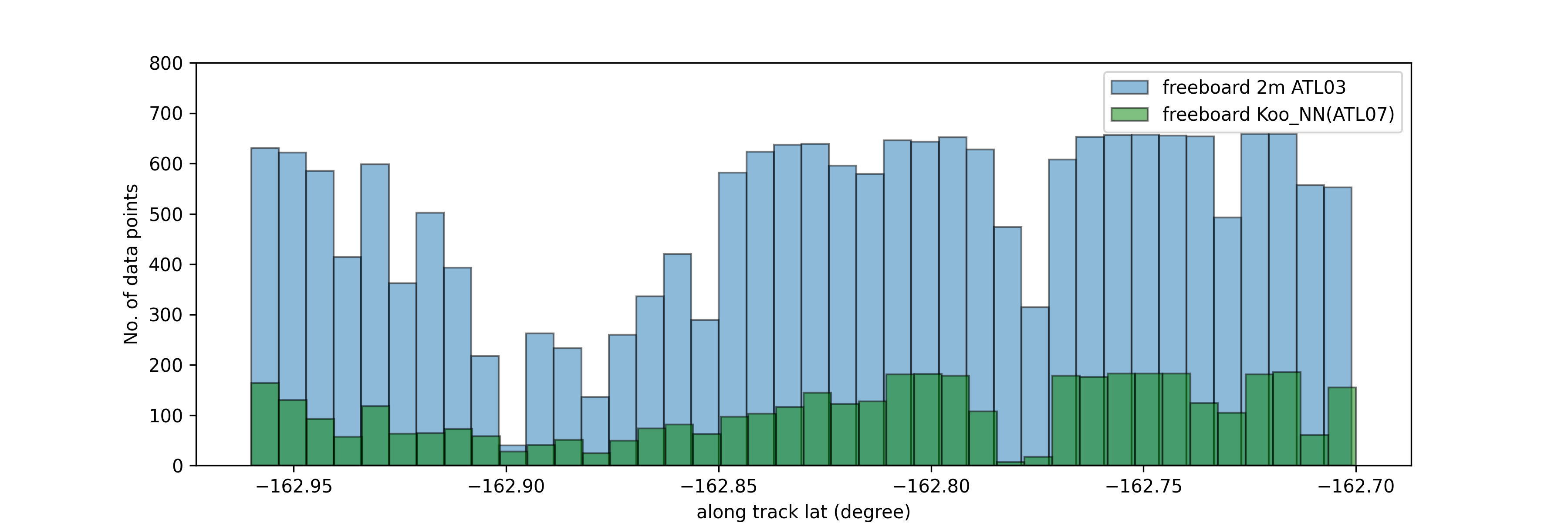}
            \caption{Point density difference between ATL03 and ATL07 (Koo\_method)}
            \label{fig:hist26_atl_03_10}
        \end{subfigure}
    \caption{Freeboard comparison, (a),(b) represents ATL03 and ATL07/ATL10 freeboard, (c) freeboard value density ATL03 with ATL07/ATL10, and (d) represent point density difference between ATL03 with ATL07/ATL10 freeboard over IS2 track \textit{20191126182014\_09290510\_gt2r}}
    \label{fig:freeboard_comp2}
\end{figure}

\subsubsection{Freeboard Comparison}
The figures \ref{fig:freeboard_comp1} and \ref{fig:freeboard_comp2} show the freeboard comparisons from this study on ATL03 and those from Koo\_method and ATL10. 
Clearly, our results directly based on 2m sampled ATL03 data show a more dense and high-resolution freeboard product than those based on the ATL07 and ATL10 products, although the freeboard distributions show similar peak values. 

\subsubsection{Freeboard Computation Speedup}
Similar to the auto-labeling process, we utilize a PySpark-based MapReduce framework to scale the freeboard computation. The freeboard calculation for IS2 ATL03 data benefits from high scalability due to the independent processing of data points. PySpark enables the parallelization and scaling of the freeboard calculation across different architectures. 
\begin{table}[htb]
\centering
\caption{PySpark-based IS2 freeboard computation over Google Cloud.}
\label{tab:freeboard_spark}
\begin{tabular}{ |>{\centering\arraybackslash}p{0.115\linewidth}||>{\centering\arraybackslash}p{0.06\linewidth}||>{\centering\arraybackslash}p{0.06\linewidth}||>{\centering\arraybackslash}p{0.06\linewidth}||>{\centering\arraybackslash}p{0.08\linewidth}||>{\centering\arraybackslash}p{0.11\linewidth}||>{\centering\arraybackslash}p{0.11\linewidth}|}
\hline
\textbf{Executors} & \textbf{Cores} & \textbf{Load Time (s)} & \textbf{Map Time (s)} & \textbf{Reduce Time (s)} & \textbf{Speed-up Load} & \textbf{Speed-up Reduce} \\ \hline
1                  & 1              & 111                & 0.4               & 392                  & 1                      & 1                        \\ \hline
1                  & 2              & 60                 & 0.4               & 177                  & 1.85                   & 2.21                     \\ \hline
1                  & 4              & 36                 & 0.3               & 74                   & 3.08                   & 5.30                     \\ \hline
2                  & 1              & 58                 & 0.3               & 159                  & 1.91                   & 2.47                      \\ \hline
2                  & 2              & 33                 & 0.3               & 86                   & 3.36                   & 4.56                     \\ \hline
2                  & 4              & 21                 & 0.3               & 44                   & 5.29                   & 8.91                     \\ \hline
4                  & 1              & 34                 & 0.2               & 80                   & 3.26                   & 4.9                        \\ \hline
4                  & 2              & 20                 & 0.2               & 41                   & 5.55                   & 9.56                       \\ \hline
4                  & 4              & 13                 & 0.3               & 25                   & \textbf{8.54}                      & \textbf{15.68}                    \\ \hline
\end{tabular}
\end{table}
This includes not only single multi-core machines but also multiple heterogeneous machines within a GCD cluster, achieving an 8.5x improvement in data loading and a 15.7x speedup in map-reduce processing as displayed in Table \ref{tab:freeboard_spark}. The PySpark-based approach's support for larger clusters enhances the potential for scaling the freeboard computation to much larger datasets in the future.

%% file: sections/conclusion.tex
This research explores the ICESat-2 ATL03 2m sampled data for sea ice classification and freeboard retrieval. We automatically labeled sea ice on ATL03 data using correlated color-based-thin-cloud-shadow-filtered labeled S2 imagery for training and scaled the process.
Our sea ice classification results on ATL03 data indicate that the LSTM model provides more accurate results in classifying thick ice, thin ice, and open water in the polar regions than the MLP model. We successfully scaled and distributed the deep learning training over multiple GPUs using the Horovod framework and achieved a better speedup. 
We also achieved a better resolution of the local sea surface height. We calculated a better resolution of freeboard information along the 2m sampled ALT03 track than those based on the ATL07 data.
\paragraph{Future Work}
Still, downloading the massive ATL03 data to local computers for processing is a big challenge. The future of this work is to directly access the data from the Cloud, while combined with scaled and distributed deep learning to speedup the processing and generate polar-wide scale freeboard and even thickness products. 
In the end, better sea ice products will help domain scientists better understand sea ice dynamics and changes in a warming climate.